\documentclass[10pt,letterpaper]{article}
\usepackage[top=0.85in,left=2.75in,footskip=0.75in]{geometry}

\usepackage{amsmath,amssymb}

\usepackage{changepage}

\usepackage{textcomp,marvosym}

\usepackage{nameref,hyperref}

\usepackage[right]{lineno}

\usepackage[nopatch=eqnum]{microtype}
\DisableLigatures[f]{encoding = *, family = * }

\usepackage[table]{xcolor}
\usepackage{times}

\usepackage{array}
\usepackage{cite}
\usepackage{amsmath}
\usepackage{booktabs}
\usepackage{cleveref}
\usepackage{tabularx} 
\usepackage{multirow}
\usepackage{adjustbox}
\usepackage{graphicx}
\usepackage{subcaption}
\newcolumntype{+}{!{\vrule width 2pt}}

\newlength\savedwidth

\usepackage[aboveskip=1pt,labelfont=bf,labelsep=period,justification=raggedright,singlelinecheck=off]{caption}

\makeatletter
\renewcommand{\@biblabel}[1]{\quad#1.}
\makeatother

\usepackage{lastpage,fancyhdr,graphicx}
\usepackage{epstopdf}
\fancyheadoffset[L]{2.25in}
\fancyfootoffset[L]{2.25in}
\begin{document}
\vspace*{0.2in}

\begin{flushleft}
{\Large
\textbf{Explainable Emotion and Mental Health Reasoning in Large Language Models Using Theories of Emotion} 
}
\newline
\\
Tadesse Destaw Belay\textsuperscript{1}, 
Iqra Ameer\textsuperscript{2},
Helena Gómez Adorno\textsuperscript{3},
Alexander Gelbukh\textsuperscript{1}, \\
Eusebio Ricardez Vázquez\textsuperscript{1},
Olga Kolesnikova\textsuperscript{1},
Grigori Sidorov\textsuperscript{1},
Seid Muhie Yimam\textsuperscript{4}
\\
\bigskip

\textbf{1} Instituto Politécnico Nacional, Mexico City, Mexico \\
\textbf{2} Pennsylvania State University, Pennsylvania, USA \\
\textbf{3} Universidad Nacional Autónoma de México, Mexico City, Mexico \\
\textbf{4} University of Hamburg, Hamburg, Germany
\bigskip

%
%









\end{flushleft}


%

\section*{Abstract}
Advancements in Large Language Models (LLMs) have enabled a wide range of applications. However, their ability to simulate human-like behavior and reasoning remains unclear. While emotion detection is a better-established Natural Language Processing (NLP) task, existing approaches primarily focus on directly classifying text into predefined emotion classes, often overlooking the appraisal process that explains why an individual experiences a particular emotion in a given situation.
In this work, we propose Psychology-based Chain-of-Thought (PsyCoT) reasoning, a theory-driven prompting strategy grounded in the Schachter–Singer and Cognitive Appraisal theories of emotion. We propose and evaluate PsyCoT for subjective NLP tasks, including emotion analysis, emotion-intensity prediction, stress analysis, and depression-severity identification. Our evaluation covers four levels of classification complexity: binary, multiclass, multi-label, and ordinal scaling, spanning 11 languages. We assess the performance of five state-of-the-art LLMs across these selected subjective NLP tasks.
Experimental results demonstrate that PsyCoT prompt variants improve the standard chain-of-thought (CoT) prompting across the evaluated tasks. Specifically, in multi-label emotion classification, PsyCoT yields F1 score gains ranging from 0.5\% to 28.1\%, depending on the model and language configuration. For multi-label emotion intensity prediction, the Schachter-Singer Two-Factor Theory (2FT) PsyCoT variant shows a clear advantage over both standard CoT and Cognitive Appraisal Theory (CAT) prompting, achieving overall improvements across models with gains up to +4.8\%. For stress prediction, the 2FT PsyCoT variant consistently outperforms standard CoT and CAT prompting across evaluated models, achieving substantial F1 score gains (up to +14.6\%, such as Gemma-3-27B rising to 76.5\% F1 compared to 63.1\% for CoT). Conversely, for depression severity, the CAT variant tends to show a stronger advantage, yielding the highest F1 scores across most models. These findings highlight the effectiveness of the proposed psychology-grounded reasoning in improving both explainability and performance in emotion and mental-health NLP tasks.



\section{Introduction}

Large Language Models (LLMs) have demonstrated significant progress in the evaluation of downstream Natural Language Processing (NLP) tasks, enabling human-like language understanding and text generation across diverse domains, including education \cite{chu-etal-2025-llm,movassagh2023artificial}, healthcare applications \cite{maity2025large}, and mental health support \cite{10.1145/3544548.3580688}. In parallel, prompt engineering techniques have become essential for maximizing LLMs' capabilities \cite{ramnath-etal-2025-systematic,vatsal2024surveypromptengineeringmethods}.
Chain-of-Thought (CoT) reasoning has shown strong performance gains in tasks requiring multi-step inference, particularly in mathematical, symbolic, and commonsense reasoning \cite{wei2022chain,10.1145/3774896}. These developments position LLMs as promising tools for tackling complex NLP tasks.

One such complex and subjective area of NLP is affective computing tasks \cite{zhang2025affectivecomputingeralarge}. The affective computing field has progressed beyond coarse-grained sentiment analysis (positive/negative) toward fine-grained emotion recognition \cite{mohammad-etal-2018-semeval}, emotion intensity estimation \cite{muhammad-etal-2025-semeval}, and mental health status assessment \cite{sharma2025assessment}. The current open-weight and closed-source LLM families, such as OpenAI’s GPT, Google’s Gemma and Gemini, Meta’s LLaMA, etc., have shown encouraging contextual learning (zero-shot and few-shot performance) on these subjective NLP tasks, suggesting that they internalize rich and affective knowledge during pretraining \cite{song2025largelanguagemodelssubjective}. As a result, the research has increasingly explored prompt-based and instruction-tuned approaches for emotion analysis and mental health–related inferences \cite{emollms}.

However, as LLMs are deployed in high-stakes affective applications, such as detecting depressive symptoms \cite{weber2025using}, psychological distress \cite{SETTANNI2025116583}, or suicide risk \cite{holmes2025applications}, the emphasis shifts from classification accuracy alone to the explainability of the reasoning behind the decision. In this setting, it is insufficient for a model to label a text as sad, stressed, or anxious; it should also provide a rationale for these labels that is coherent, transparent, and psychologically meaningful. Reasoning that lacks grounding in human emotion processes risks being misleading, uninformative, or clinically unusable \cite{crivelli2024emotional}. From a cognitive perspective, subjective language refers to the speaker’s or writer’s internal states, such as emotions, attitudes, beliefs, or intentions. Cognitive and psychological studies of language indicate that understanding subjectivity often requires a theory of mind (i.e., inferring the speaker’s intent or feelings) and empathy. For instance, sarcasm and irony contents are typical subjective uses of language: the literal content diverges from the intended meaning, and the listener must infer the speaker’s true attitude \cite{song2025largelanguagemodelssubjective}.

To address the interpretability of subjective NLP tasks, most existing approaches rely on Chain-of-Thought (CoT) prompting \cite{wei2022chain}, which encourages models to produce intermediate reasoning steps. While CoT is effective for mathematical and symbolic reasoning tasks governed by formal logic, it has a critical limitation when applied to psychology-oriented NLP tasks. This is because human emotion is not the result of linear logical deduction; rather, it emerges from complex interactions among situational stimuli, cognitive interpretation, and physiological mental states \cite{vsimic2021understanding}. Consequently, the generic CoT principle of \textit{“think step-by-step”} reasoning often yields explanations that are linguistically plausible but psychologically shallow. This raises a fundamental question: \textit{What does “think step-by-step” actually mean in the context of human emotion and mental health state analysis?}. This is because emotion analysis is inherently subjective (varying across individuals, cultures, languages, and situations) and context-dependent \cite{JIM2024100059}. Psychological research emphasizes that emotions are processes (not labels) shaped by appraisal, arousal, and interpretation \cite{yeo2024associations}. The standard CoT prompting largely ignores this psychological grounding, treating affective inference as a generic reasoning problem. As a result, LLMs may simulate emotional outcomes without modeling the mechanisms that generate them. 

Human cognitive processes have been studied in numerous psychological theories of emotion. Schachter–Singer Two-Factor Theory (2FT)\cite{schachter1962cognitive} and Cognitive Appraisal Theory (CAT) are among the most common and well-known emotion theories \cite{lazarus1991progress}. The 2FT theory posits that emotion arises from the interaction between physiological arousal and cognitive labeling of that arousal. Conversely, CAT argues that emotions are elicited by an individual’s evaluation of events in relation to personal goals, beliefs, and well-being \cite{lazarus1991progress}.

Achieving meaningful explainability in affective computing requires that LLM reasoning be grounded in psychological theory rather than generic logical decomposition \cite{DONG2025101014}. This shift moves from prompting LLMs to “think like logicians” toward prompting them to “process emotions like humans.” To this end, we propose a theory-driven prompting framework that operationalizes two established psychological theories of emotion into structured prompting schemas for affective computing NLP tasks: the \textbf{Schachter–Singer Two-Factor Theory (2FT)}, which characterizes emotion as a combination of physiological arousal and cognitive labeling \cite{schachter1962cognitive}, and the \textbf{Cognitive Appraisal Theory (CAT)}, which posits that emotions arise from individuals’ evaluations of events in relation to their goals and well-being \cite{lazarus1991progress}.

These theories suggest that understanding why an emotion occurs requires analyzing antecedent events, appraisal dimensions, and arousal-related components—elements that are not explicitly represented in standard CoT prompting. The proposed theory-based prompts guide LLMs to (1) identify situational antecedents, (2) perform cognitive appraisals, (3) infer physiological or arousal-related indicators, and (4) integrate these components into coherent inferences about emotion category, emotional intensity, and mental health status (e.g., stress and depression severity). 
This work tends to answer the following Research questions (RQs):
\begin{itemize}
    \item \textit{RQ1:} Can LLMs mimic human psychological reasoning in assessing and explaining emotional and mental health states?
    \item \textit{RQ2:} How does psychologically grounded reasoning improve LLMs' emotion understanding compared to chain-of-thought (COT) reasoning?
    \item \textit{RQ3:} How does the emotional intelligence of LLMs vary across different prompt variants, model architectures, parameter sizes, languages, and affective computing NLP tasks?
\end{itemize}

In summary, this work introduces a \textit{Psychology-based Chain-of-Thought (PsyCoT)} prompting strategy grounded in the Schachter–Singer and Cognitive Appraisal theories of emotion. PsyCoT embodies a shift from prompting LLMs to ``think like logicians'' toward prompting them to ``process emotions like humans." By embedding established theories of emotion into LLM prompting, we aim to bridge the gap between high-performing affective inference and psychologically grounded explainability, thereby enhancing the reliability and practical applicability of LLMs in emotion and mental health analysis.

The rest of this paper is organized as follows. Section \ref{sec:related} reviews related work on prompting strategies for emotion and health-related NLP tasks. Section \ref{sec:theory} presents the theoretical foundations of the proposed psychology-based Chain-of-Thought (PsyCoT) framework for modeling emotion. Section \ref{sec:evalaute} describes the evaluation tasks and dataset details. Section \ref{sec:experiemnt} outlines the experimental setup, including prompts, selected large language models (LLMs), and evaluation parameters. Section \ref{sec:results} reports the findings, along with result analysis and discussion, and the paper concludes in Section \ref{sec:conclusion}.

\section{Related works} \label{sec:related}
Advancing the performance of LLMs for subjective NLP tasks has various applications, from personalizing content or user recommendations in social networks \cite{10970300} to empathetic dialogue systems \cite{bai2025holistic}. In subjective NLP tasks, no single correct answer may exist; instead, better reasoning is a key milestone toward modeling artificial general intelligence (AGI) \cite{bhattacharyya2025do}. Several approaches have been developed to enhance LLMs’ emotional capabilities, including in-context learning \cite{dong-etal-2024-survey}, language model fine-tuning methods \cite{PRATAP2025100144}, and prompting approaches \cite{li2024enhancing}. However, challenges remain, including overreliance on annotated data, difficulty handling complex emotions, and the difficulty of explaining LLMs' decision-making in emotion cognition. Moreover, processing human emotions requires not only computational methods and techniques but also a deep understanding of how humans perceive emotion (psychological theories) \cite{chen2024recentadvancement}.

In the past decade, the explainability of deep learning approaches has advanced rapidly, ranging from explainable deep learning frameworks for prediction, such as DeepHINT \cite{bty842}, to methods for interpreting generative language models, such as Captum \cite{miglani-etal-2023-using}.  
At the same time, general-purpose language models have demonstrated strong performance across a wide range of tasks, including simulating human-like behaviors such as emotional intelligence. As users interact with generative models via prompts or instructions, various prompting strategies have been proposed for evaluating downstream NLP tasks. In this related work section, we discuss the most common prompting approaches for assessing LLMs' understanding of affective phenomena.

\subsection{Standard (regular) prompt}
Standard prompting is a simple approach in which a language model is given a direct question, with or without context examples of input–output pairs before generating a predicted answer \cite{fewshot-learners}. In this setting, the model directly produces an answer, without step-by-step processing of the given tasks. For example, in a multi-label emotion identification task, the prompt could look like  \textit{``Categorize the text’s emotional tone as the presence of none, one or more of the given emotions \{emotion classes list here ...}"\} \cite{belay-etal-2025-evaluating}. However, the standard prompting approach is outperformed by the “think step-by-step” chain-of-thought (CoT) prompting strategies in multiple downstream NLP tasks \cite{wei2022chain}.

\subsection{Chain-of-Thought (CoT) prompt}
To leverage LLMs' reasoning capabilities, \cite{wei2022chain} proposed Chain-of-Thought (CoT) prompting, which appends multiple reasoning steps before answering the input question. With this simple prompting strategy, LLMs perform substantially better on complex problems and reasoning tasks such as mathematical problems \cite{NEURIPS2022_18abbeef} and commonsense reasoning \cite{chae-etal-2023-dialogue}. The CoT prompting approach is commonly appended with \textit{``think step-by-step"} in standard prompting strategies. However, CoT does not consistently perform well across all NLP tasks, and performance varies with downstream evaluation tasks \cite{Barez-undated-xk}. In this case, cognitive science prompting (emotional prompt) approaches aim to reduce specific failure modes in the processing of subjective tasks, such as by adding a short emotional stimulus to standard and/or CoT prompts \cite{li2023large}.

\subsection{Emotional prompts}

The work \cite{li2023large} shows that LLMs exhibit emotional intelligence, and the performance can be improved by combining standard prompts with emotional stimuli. Emotional stimuli are psychological phrases that come after the standard or CoT prompts, such as \textit{``This is very important to my career."}, \textit{``Are you sure?"}, and other related phrases/sentences. The work \cite{li2024ECOT} proposed Emotional Chain-of-Thought (ECoT), which inserts three manually crafted reasoning steps into the conversation to understand emotions before producing an answer. These steps typically include: (1) identifying the listener's emotions and explaining why, (2) identifying the speaker's emotions and explaining why, and (3) formulating an empathetic reply to the listener and other involved parties. ECoT prompts yield promising results for emotion recognition compared to standard and CoT prompting approaches. However, these steps are handcrafted and task-dependent, and LLMs may misinterpret emotional cues in the prompts, leading to inaccurate or inappropriate responses.

Chain of Empathy (CoE) prompting \cite{chain-of-emphathy} uses insights from psychotherapy to prompt LLMs to reason about human emotional states. The common reasoning steps involved in CoE are: (1) Identify any word that represents the client’s emotion, and (2) Understand the individual/situational factors that may have led to the expression in the client’s message. It aims to understand and respond to human emotions by mimicking therapeutic reasoning, using frameworks from psychotherapy such as Cognitive-Behavioral Therapy (CBT) \cite{beck1979cognitive}, Dialectical Behavior Therapy (DBT) \cite{linehan1987dialectical}, Person-Centered Therapy (PCT) \cite{cooper2011person}, and Reality Therapy (RT) \cite{Glasser1964} to analyze a user's feelings. These psychotherapy frameworks address underlying causes and context, and lead to more nuanced, relevant, and genuinely empathetic interactions \cite{chainofempathyrl}.

\subsection{Cognitive appraisal-based prompts}
Cognitive appraisal theories of emotion assert that emotions stem from an individual’s subjective understanding and interpretation of the situation \cite{ortony2022cognitive,zhan-etal-2023-evaluating,yeo-jaidka-2025-beyond}. Cognitive appraisal-based emotion datasets are created using LLMs \cite{zhan-etal-2023-evaluating} and studies such as \cite{yeo-jaidka-2023-peace,li2024enhancing,yeo-jaidka-2025-beyond} elicit individuals' emotional and appraisal experiences during interactions with personally significant products or services. The work \cite{croissant2024appraisal} proposed appraisal-based Chain-of-Emotion prompting for conversational affective game agents. However, cognitive appraisal-based reasoning has not yet been explored for existing emotion and mental health datasets. Based on this, we propose emotion theory-based prompts to enhance LLMs' ability to analyze emotional and mental health states.
\begin{figure*}[!ht]
\begin{adjustwidth}{-2.25in}{0in}

    \centering
    \includegraphics[width=1\linewidth]{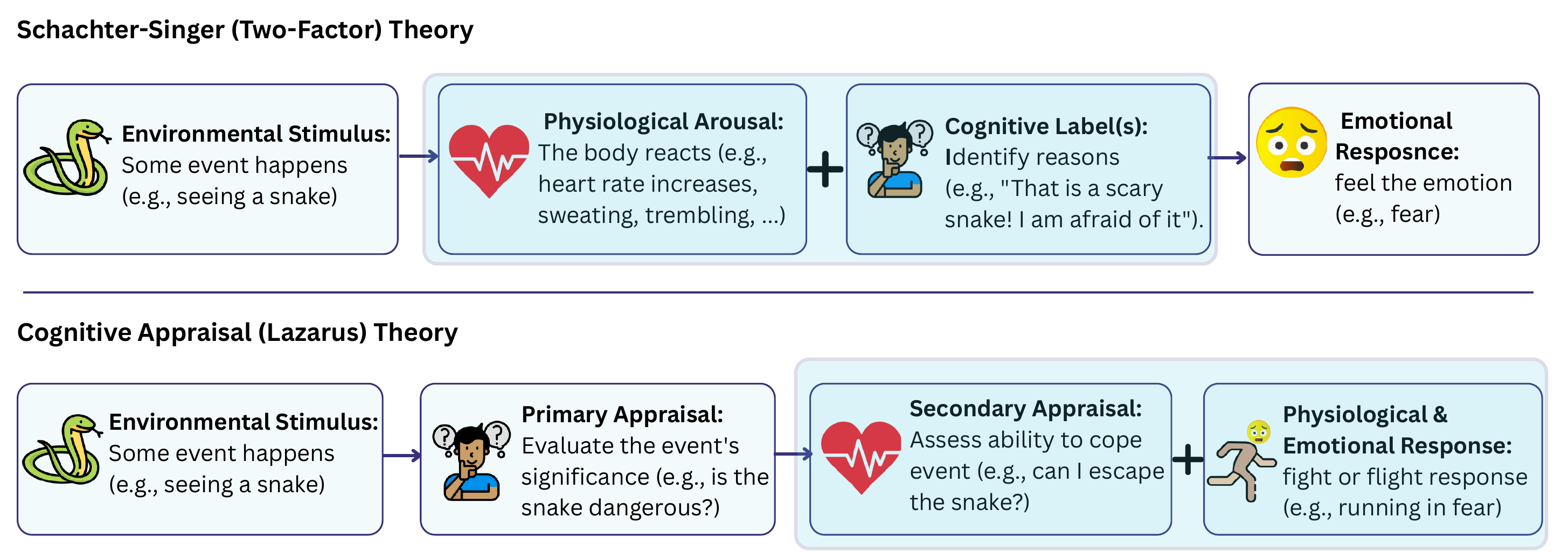}
    \caption{The steps in experiencing emotion in the Schachter-Singer Two-Factor and Cognitive Appraisal theories of emotion in response to the same environmental stimuli. Actions shown within the inside box occur simultaneously for the corresponding theory of emotion.}
    \label{fig:theories}
\end{adjustwidth}
\end{figure*}

\section{Theoretical frameworks of emotion}\label{sec:theory}
Emotions are fundamental to human survival and behavior, driving our actions and experiences and giving life meaning, which underscores the importance of understanding them \cite{SMITH2020107925, ekman1992argument}. Although emotion detection is a better-established NLP task \cite{mohammad-etal-2018-semeval,belay-etal-2025-evaluating,muhammad-etal-2025-semeval}, it mostly involves classifying text directly into emotion labels, skipping the appraisal step necessary to interpret why an individual experiences an emotion in a particular event.
The following are two well-known theories of emotion that explain how humans perceive emotions.

\subsection{Schachter-Singer (two-factor) theory}
The Schachter-Singer Two-Factor Theory (2FT) was developed by \cite{schachter1962cognitive}. It proposes that emotion arises from the combination of physiological arousal and a cognitive evaluation of bodily responses. According to this theory, when an event causes physiological arousal, we interpret or label that arousal in light of the situation, and this interpretation leads to the emotion. The two factors are 1) \textbf{Arousal} - the physical reaction to a stimulus (e.g., sweating, rapid breathing, or a racing heart) and 2) \textbf{Cognitive label} - identifying a reason for the arousal or cognitive interpretation of this arousal (e.g., “I am scared because there is a snake”). 

For example, when a person encounters a snake, their body exhibits physiological arousal, such as an increased heart rate. This arousal is then cognitively interpreted as fear because the situation is perceived as dangerous. According to this theory, emotion arises from the combination of physiological arousal and its cognitive interpretation. The theory further suggests that similar physiological responses can lead to different emotions depending on the context. For instance, a racing heart and sweaty palms during an exam may be interpreted as anxiety, whereas the same responses on a date may be perceived as affection, attraction, or excitement, as illustrated in Fig. \ref{fig:theories}.

\subsection{Cognitive appraisal theory}

The Cognitive Appraisal Theory (CAT), pioneered by \cite{lazarus1991progress}, proposes that cognitive evaluation precedes emotional experience. In this view of the theory, a person first encounters a stimulus, then interprets it, and simultaneously experiences both the physiological response and the emotion. Cognitive appraisal typically involves two stages. In the primary appraisal, individuals evaluate the significance of an event for their well-being (e.g., “Is the snake dangerous?”). In the secondary appraisal, they assess their ability to cope with the event (e.g., ``Can I escape the snake?”).

According to cognitive appraisal theories, the basic premise is that events themselves do not directly cause emotions and behaviors; rather, they do so through how those events are perceived or appraised \cite{lazarus1991progress}. Consequently, different interpretations of the same eliciting event can lead different people to experience different emotions. Cognitive Appraisal Theory helps explain why people can react differently to the same situation \cite{appraisal-Moors2017}. 
For example, encountering a snake may be immediately appraised as a serious threat, triggering fear and eliciting the fight-or-flight response, as shown in Fig \ref{fig:theories}.

Both Schachter–Singer and Cognitive Appraisal theories rely on interpreting or labeling internal states. However, in the Schachter–Singer theory, emotion arises from the simultaneous interaction of physiological arousal and this interpretive process. In contrast, Cognitive Appraisal theory proposes that individuals first evaluate and label the stimulus, after which physiological and emotional responses follow together based on that appraisal. Inspired by prominent emotion theories, we proposed Psychology-based Chain-of-Thought (PsyCoT) prompts grounded in the Schachter–Singer Theory and the Cognitive Appraisal Theory to improve LLMs' reasoning for subjective NLP tasks.  Unlike approaches that aim to directly simulate human emotional experiences, PsyCoT employs established psychological theories as structured reasoning guidelines to encourage LLMs to consider contextual antecedents, physiological arousal, and cognitive appraisals when predicting emotional states. The resulting explanations are therefore intended to provide theory-informed rationales for model predictions, rather than validated representations of human psychological processes or the models' internal reasoning mechanisms.

\section{Evaluation tasks and datasets}\label{sec:evalaute}
For the proposed Psychology-based Chain-of-Thought (PsyCoT) prompts, we select evaluation tasks that involve subjectivity and different levels of reasoning complexity. The tasks include binary classification (e.g., stress/not-stress), multi-label and multi-class classification (e.g., multi-label emotion analysis, depression severity levels), and scaling tasks (e.g., emotion intensity scale: no emotion, low, medium, high). We examine the proposed physiology-based emotion prompting across four tasks and datasets. The details of the selected evaluation tasks, along with data examples, are presented in the following sections, with a summary in Table \ref{tab:example}.

\begin{table}[!h]
\begin{adjustwidth}{-2.25in}{0in}
\centering
\caption{\textbf{Evaluation tasks, data type details, and an example from the English language (the example texts are written as they appear in the original data). We select a 1K random sample test set for each language if the language test set has more than 1K samples.}}

\resizebox{\dimexpr\textwidth+2.25in}{!}{
\begin{tabular}{p{1cm}p{6.5cm}p{7cm}p{1.9cm}p{1.7cm}}
\toprule
\textbf{Dataset} & \textbf{Task type} & \textbf{Example} & \textbf{Label(s)} & \textbf{Test size} \\
\midrule

\cite{muhammad-etal-2025-brighter,belay-etal-2025-evaluating} &
\textbf{Multi-label emotion analysis}: [``anger", ``disgust", ``fear", ``joy", ``sadness", ``surprise"], 
\textbf{Languages}: [amh, arq, chn, deu, eng, esp, hau, ptbr, ron, rus, ukr] &
``back in 2006 when i was in high school i took earth sciences, and in the spring i was on my teacher's computer when i came across a photo of me giving the thumbs up." &
\shortstack{[``joy",\\ ``surprise"]} &
1K for each language \\

\midrule

\cite{muhammad-etal-2025-semeval} &
\textbf{Emotion intensity analysis}: [0, 1(low), 2(medium), 3(high)], 
\textbf{Languages}: [amh, arq, chn, deu, eng, esp, hau, ptbr, ron, rus, ukr] &
``idk why, my eyes just got really watery and I felt really scared and felt so lost all of a sudden:" &
[``fear":3, ``sadness":2,``surprise":1] &
1K for each language \\

\midrule

\cite{turcan-mckeown-2019-dreaddit} &
\textbf{Binary stress prediction}: [``stress", ``not stress"], 
\textbf{Language}: [eng] &
``I was uncomfortable. He’d alleviate my fears. We’d progress. Then it was something new. Again, I was uncomfortable." &
stress &
1K \\

\midrule

\cite{depseverty} &
\textbf{Depression severity prediction}: [``minimum", ``mild", ``moderate", ``severe"],
\textbf{Language}: [eng] &
``Idk Do I tell someone? Do I just quit? Do I talk to her about what she did? Please, any advice would be really really helpful to me!" &
severe &
716 \\

\bottomrule
\end{tabular}
}
\label{tab:example}
\end{adjustwidth}
\end{table}

\subsection{Multi-label emotion analysis}
Multi-label emotion classification aims to predict all emotions expressed in a text rather than assigning a single label. We use the SemEval-2025 Task 11 dataset \cite{muhammad-etal-2025-brighter,muhammad-etal-2025-semeval}, which comprises 32 languages and is annotated with multi-label emotion categories and associated intensity scores for six basic emotions (anger, disgust, fear, joy, sadness, and surprise)\footnote{\url{https://huggingface.co/datasets/brighter-dataset/BRIGHTER-emotion-intensities}}. From this dataset, we select 11 languages based on the following criteria.
\begin{enumerate}
    \item To account for differences in language resource availability, we include low-resource languages such as Amharic (amh) and Hausa (hau); medium-resource languages including Ukrainian (ukr), Romanian (ron), Brazilian Portuguese (ptbr), and Algerian Arabic (arq); and high-resource languages such as English (eng), German (deu), Spanish (esp), Chinese/Mandarin (chn), and Russian (rus), following common language resource classifications in prior work \cite{losttranslationlargelanguage}.
    \item To capture script diversity, the selected languages span the Latin script (eng, deu, esp, ron, ptbr, hau), the Cyrillic script (ukr, rus), the Ge'ez script (amh), the Arabic script (arq), and the logographic script (chn).
    \item We restrict our selection to languages for which emotion-intensity annotations are available, thereby enabling joint evaluation of emotion classification and intensity prediction.
    \item We include only languages for which each instance is annotated by at least 5 annotators to ensure greater reliability and consistency of annotations.
\end{enumerate}

\paragraph{Task definition:} In the multi-label emotion identification task, the input is a text instance $T$, the objective is to identify all available emotion categories expressed in $T$ from a predefined six basic emotion $E$ (anger, disgust, fear, joy, sadness, and surprise) sets $E = [e1, e2, ..., e6]$. Unlike single-label classification, each text instance may be associated with none (no emotion or neutral), one, two, or multiple emotions simultaneously. The output is therefore a subset $E_T \subseteq E$ representing all emotions conveyed in the input text.

\subsection{Emotion intensity prediction} 
We use the SemEval-2025 Task 11 dataset\footnote{\url{https://huggingface.co/datasets/brighter-dataset/BRIGHTER-emotion-categories}} for emotion intensity prediction \cite{muhammad-etal-2025-semeval}. \paragraph{Task definition:} In the multi-label emotion intensity scale identification task, the input is a text instance $T$, which may be a sentence, an utterance, or a document. Given a predefined set of basic emotion categories $E = \{e1, e2, ..., e6\}$, the objective is to identify the intensity scale of each emotion expressed in $T$ using a discrete ordinal scale $S = \{0, 1, 2, 3\}$, where 0 indicates the absence of the emotion, 1 is low intensity for the referred emotion, 2 is medium intensity and 3 denotes high intensity for the corresponding emotion. The task is formulated as a multi-label problem in which multiple emotions may co-occur within a single text instance, each with an associated intensity value. The output is represented as a set of emotion–intensity pairs ${O}_T = \{ (e_i : s_i) \mid e_i \in E,\ s_i \in S \}$.

\begin{figure*}[!ht]
\begin{adjustwidth}{-2.25in}{0in}

    \centering
    \includegraphics[width=\linewidth]{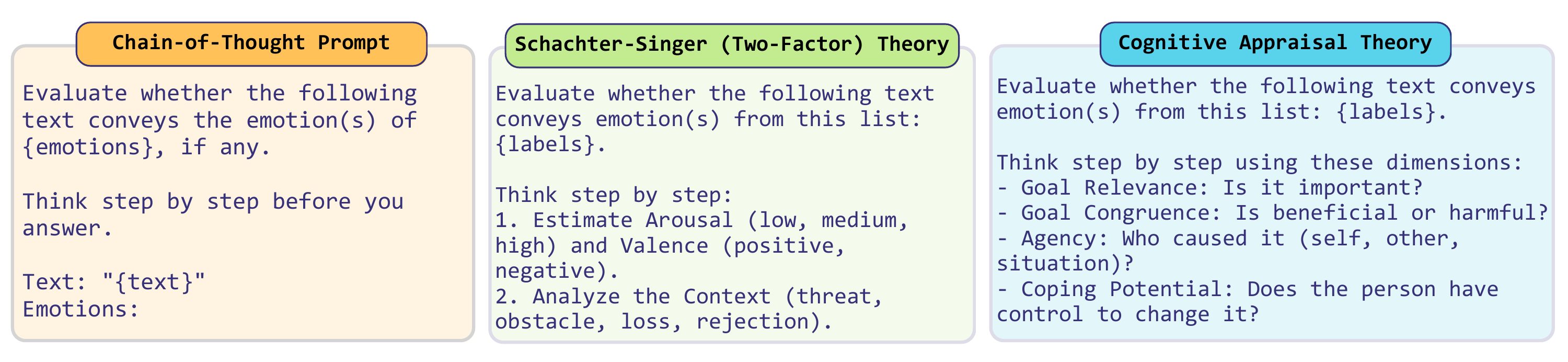}
    \caption{\textbf{The theory-based prompts used in this work. Chain-of-thought (CoT) prompting (our baseline), the Schachter-Singer (Two-Factor) Theory, and the Cognitive Appraisal Theory prompts, respectively.}}
    \label{fig:prompts}
\end{adjustwidth}
\end{figure*}

\subsection{Mental health analysis}
Mental health conditions such as depression and suicidal ideation pose serious challenges to global health care \cite{bi-etal-2025-magi}. Our experiment is based on two well-established, human-labeled datasets commonly used for mental health analysis. 
 
\paragraph{ Stress prediction:} Stress is a human experience, particularly in online environments. We select the \textbf{Dreaddit} dataset\footnote{\url{https://huggingface.co/datasets/andreagasparini/dreaddit}}, a dataset of posts collected from Reddit across five domains (abuse, social, anxiety, stress, and financial) \cite{turcan-mckeown-2019-dreaddit}. Multiple human annotators rated whether sentence segments exhibited stress or non-stress in the post, and the annotations were aggregated to produce final labels. We used this dataset to perform post-level binary stress prediction with state-of-the-art LLMs.

\paragraph{Depression severity:}
Advances in NLP systems, particularly large language models (LLMs), have played an important role in analyzing user-generated content on social media to identify risk indicators in online behavior \cite{SUPRIYONO2024100173}. We use the Depression Severity (\textbf{DepSeverity}) dataset\footnote{\url{https://github.com/KUAS-ubicomp-lab/Depression_Severity_Levels_Dataset}}, which reflects the ordinal nature of depression severity levels and is grounded in clinical depression standards applied to Reddit posts \cite{depseverty}. The posts are categorized into four levels of depression severity—minimal, mild, moderate, and severe—representing increasing intensity.

For the binary stress classification task, the input is a text instance $T$. The objective is to determine whether the text expresses stress by assigning a single label from the set $S$ \textit{{stress, not stress}}. In the depression severity classification task, the goal is to assign a single label from an ordinal set $S$ \textit{{minimal, mild, moderate, severe}}, representing increasing levels of depression severity expressed in the text.

\section{Experimental setup}\label{sec:experiemnt}
\subsection{Prompt variants}
We evaluate state-of-the-art LLMs using three prompting strategies: CoT as a baseline and two proposed Psychology-based Chain-of-Thought (PsyCoT) prompts. The prompt variants used in this work are shown in Fig \ref{fig:prompts}.

\paragraph{Chain-of-Thought (CoT) prompt:} 
 Chain-of-thought (CoT) prompting \cite{wei2022chain} achieves state-of-the-art results from both open-weight and proprietary LLMs in multidisciplinary downstream NLP tasks such as emotion analysis \cite{muhammad-etal-2025-semeval}. We use CoT prompting as our baseline and compare it to our proposed PsyCoT prompts.

\paragraph{Schachter-Singer theory-based prompt (2FT):}
Based on the Two-Factor Theory of emotion \cite{schachter1962cognitive}, this prompt instructs the LLM to decouple emotion detection into two distinct analytical steps: (1) estimating ``physiological arousal" (intensity/energy) from linguistic markers such as punctuation, capitalization, and modifiers, and (2) applying a cognitive label to this arousal based on the situational context. By explicitly isolating the arousal factor, this prompt aims to prevent the model from conflating an event's semantic severity with the emotional intensity expressed in the text, thereby improving the precision of intensity regression tasks.

\paragraph{Cognitive appraisal theory-based prompt (CAT):}
This prompt directs the LLM to simulate the cognitive evaluation process that precedes an emotional response. Instead of relying solely on semantic associations, the model is instructed to analyze the text across specific appraisal dimensions, including goal relevance, goal conduciveness, agency, and coping potential, before assigning a label \cite{lazarus1991progress}. This structured reasoning framework is designed to enhance multi-label classification by enabling the model to distinguish between emotions that share similar valence but differ in their causal structure (e.g., distinguishing \textit{Anger} from \textit{Fear} based on the presence of control).

\subsection{Large Language Models (LLMs)} 
We select the latest state-of-the-art and reasoning LLMs from both the open-weight and closed-source models based their popularity in the research community, better reasoning abilities, and from different architectures: Gemma-3-27B \cite{gemmateam2025gemma3technicalreport}, LLaMa-3.3-70B \cite{grattafiori2024llama3herdmodels}, GPT-oss-120B \cite{openai2025gptoss120bgptoss20bmodel}, Misitral-123B \cite{mistral2024large}, and 
GPT-5 \cite{singh2025openaigpt5card} from the closed-source LLMs. For ease of reproducibility and cost-effectiveness, we evaluate GPT-5-mini (Accessed: Dec 2025). All open-weight LLMs evaluated in this work are instruction-tuned (it) versions. 

\subsection{Evaluation setting and metrics}
We evaluate the LLMs in a zero-shot setting using a temperature of 0 (default), the OpenAI Chat template, and a maximum output length of 1000 tokens. For proprietary models, we use their respective APIs. The results were averaged across three runs for each model (repeating the same experiment setting three times and averaging for the final reported results). All open-weight LLMs are accessed by API keys from the open-weight LLM API providers\footnote{\url{https://gwdg.de/en/hpc/}}. 

\paragraph{Prompting LLMs:} Based on prior work showing that English prompts are superior to non-English prompts, especially for low-resource languages \cite{Barei-2024,enomoto-etal-2025-fair}, we use English prompts. For all our evaluation setups, we used 1K random samples from the test set (random-state test set data selection 42) when a language has more than 1K test samples. Given the class imbalance, we report the macro F1 score as the primary comparison metric across models and prompt variants.

\begin{table}[!b]
\begin{adjustwidth}{-2.25in}{0in}
\centering
\caption{\textbf{Multi-label emotion F1 score results using three prompt variants: The baseline chain-of-thought (CoT), the Schachter-Singer Theory (2FT), and the Cognitive Appraisal Theory (CAT) prompts. The LLMs are the instruction-tuned version (it). Results are from a zero-shot prompting approach and are averaged across the three runs.}}
\resizebox{\dimexpr\textwidth+2.25in}{!}{
\begin{tabular}{llllllllllll|l}
\toprule
\textbf{Model}&\texttt{amh}&\texttt{arq}&\texttt{chn}&\texttt{deu}&\texttt{eng}&\texttt{esp}&\texttt{hau}&\texttt{ptbr}&\texttt{ron}&\texttt{rus}&\texttt{ukr}&\textbf{Avg.} \\
\midrule
Gemma-3-27B-it-CoT & 49.6$_{\pm 0.8}$ & 15.2$_{\pm 2.4}$ & 42.7$_{\pm 3.6}$ & 45.7$_{\pm 2.5}$ & 21.2$_{\pm 4.2}$ & 20.8$_{\pm 6.3}$ & 24.2$_{\pm 5.4}$ & 29.1$_{\pm 1.3}$ & 53.0$_{\pm 1.3}$ & 24.4$_{\pm 1.9}$ & 23.4$_{\pm 6.6}$ & 31.5$_{\pm 0.7}$ \\
Gemma-3-27B-it-2FT & 49.0$_{\pm 0.8}$ & 45.3$_{\pm 0.9}$ & 54.2$_{\pm 1.1}$ & 55.5$_{\pm 0.9}$ & 46.2$_{\pm 4.1}$ & 74.9$_{\pm 1.6}$ & 55.8$_{\pm 0.5}$ & 52.7$_{\pm 1.6}$ & 62.0$_{\pm 0.2}$ & \textbf{81.4}$_{\pm 0.7}$ & \textbf{57.8}$_{\pm 0.5}$ & 57.7$_{\pm 0.5}$ \\
Gemma-3-27B-it-CAT & \textbf{50.8}$_{\pm 0.4}$ & \textbf{46.7}$_{\pm 0.6}$ & \textbf{54.6}$_{\pm 0.4}$ & \textbf{57.9}$_{\pm 0.8}$ & \textbf{56.4}$_{\pm 6.5}$ & \textbf{76.4}$_{\pm 0.1}$ & \textbf{55.8}$_{\pm 0.8}$ & \textbf{55.4}$_{\pm 0.5}$ & \textbf{64.0}$_{\pm 0.6}$ & 79.6$_{\pm 0.4}$ & 56.7$_{\pm 0.6}$ & \textbf{59.5}$_{\pm 0.7}$ \\
\midrule
LLaMa-3.3-70B-it-CoT & 38.5$_{\pm 1.0}$ & 46.0$_{\pm 0.7}$ & \textbf{52.9}$_{\pm 0.3}$ & 58.4$_{\pm 0.9}$ & 55.1$_{\pm 5.8}$ & 76.0$_{\pm 1.0}$ & 46.9$_{\pm 1.0}$ & 51.8$_{\pm 0.6}$ & \textbf{61.4}$_{\pm 0.0}$ & 77.0$_{\pm 0.8}$ & 51.9$_{\pm 0.5}$ & 56.2$_{\pm 0.3}$ \\
LLaMa-3.3-70B-it-2FT & 39.0$_{\pm 0.2}$ & 45.6$_{\pm 0.6}$ & 54.4$_{\pm 0.6}$ & 57.2$_{\pm 0.8}$ & \textbf{51.4}$_{\pm 4.8}$ & 76.1$_{\pm 0.2}$ & 44.5$_{\pm 1.1}$ & 51.4$_{\pm 0.5}$ & 60.8$_{\pm 0.5}$ & \textbf{78.7}$_{\pm 0.7}$ & 51.2$_{\pm 0.7}$ & 55.5$_{\pm 0.5}$ \\
LLaMa-3.3-70B-it-CAT & \textbf{38.6}$_{\pm 0.5}$ & \textbf{47.9}$_{\pm 0.6}$ & 52.8$_{\pm 0.7}$ & \textbf{58.5}$_{\pm 1.1}$ & \textbf{54.8}$_{\pm 6.7}$ & \textbf{76.1}$_{\pm 0.3}$ & \textbf{47.1}$_{\pm 1.2}$ & 50.9$_{\pm 0.8}$ & 63.1$_{\pm 0.3}$ & 77.5$_{\pm 0.2}$ & \textbf{52.4}$_{\pm 1.2}$ & \textbf{56.6}$_{\pm 0.7}$ \\
\midrule
Mistral-123B-it-CoT & 25.5$_{\pm 5.4}$ & 27.7$_{\pm 10.2}$ & 38.5$_{\pm 6.0}$ & 52.7$_{\pm 8.1}$ & 42.0$_{\pm 4.4}$ & 53.4$_{\pm 8.2}$ & 24.2$_{\pm 2.7}$ & 37.1$_{\pm 13.4}$ & 40.2$_{\pm 12.4}$ & 59.8$_{\pm 15.8}$ & 38.6$_{\pm 10.2}$ & 40.0$_{\pm 7.5}$ \\
Mistral-123B-it-2FT & 28.8$_{\pm 7.7}$ & 35.2$_{\pm 1.7}$ & 41.0$_{\pm 6.3}$ & 55.2$_{\pm 2.3}$ & 48.0$_{\pm 7.7}$ & 66.2$_{\pm 2.6}$ & 25.7$_{\pm 2.5}$ & 47.1$_{\pm 3.9}$ & 56.5$_{\pm 1.9}$ & 71.0$_{\pm 8.2}$ & 48.7$_{\pm 2.3}$ & 47.6$_{\pm 2.6}$ \\
Mistral-123B-it-CAT & \textbf{29.8}$_{\pm 6.9}$ & \textbf{44.4}$_{\pm 3.8}$ & \textbf{47.1}$_{\pm 3.7}$ & \textbf{58.4}$_{\pm 3.6}$ & \textbf{52.0}$_{\pm 5.9}$ & \textbf{67.2}$_{\pm 4.4}$ & \textbf{33.5}$_{\pm 3.0}$ & \textbf{53.7}$_{\pm 0.9}$ & \textbf{58.6}$_{\pm 1.5}$ & \textbf{76.1}$_{\pm 1.2}$ & \textbf{49.5}$_{\pm 2.6}$ & \textbf{51.9}$_{\pm 0.2}$ \\
\midrule
GPT-oss-120B-CoT & 42.5$_{\pm 1.2}$ & 42.1$_{\pm 2.0}$ & 55.8$_{\pm 0.8}$ & 61.8$_{\pm 2.6}$ & 53.0$_{\pm 7.4}$ & 75.7$_{\pm 1.4}$ & 51.1$_{\pm 1.1}$ & 57.1$_{\pm 0.8}$ & 61.3$_{\pm 2.5}$ & \textbf{83.7}$_{\pm 1.0}$ & 58.5$_{\pm 0.4}$ & 58.4$_{\pm 1.5}$ \\
GPT-oss-120B-2FT & 38.6$_{\pm 1.3}$ & 41.4$_{\pm 0.2}$ & 53.0$_{\pm 1.1}$ & 57.8$_{\pm 1.0}$ & \textbf{54.5}$_{\pm 6.8}$ & 72.9$_{\pm 0.2}$ & 50.3$_{\pm 0.4}$ & 54.4$_{\pm 0.5}$ & 54.6$_{\pm 0.8}$ & 82.8$_{\pm 1.0}$ & \textbf{56.9}$_{\pm 0.3}$ & 56.1$_{\pm 0.6}$ \\
GPT-oss-120B-CAT & \textbf{43.4}$_{\pm 1.3}$ & \textbf{43.9}$_{\pm 2.8}$ & \textbf{56.2}$_{\pm 0.5}$ & \textbf{63.1}$_{\pm 2.7}$ & 54.2$_{\pm 4.9}$ & \textbf{77.1}$_{\pm 2.1}$ & \textbf{51.7}$_{\pm 1.5}$ & \textbf{58.2}$_{\pm 0.4}$ & \textbf{62.2}$_{\pm 2.0}$ & 83.0$_{\pm 0.9}$ & 58.0$_{\pm 1.0}$ & \textbf{59.2}$_{\pm 0.6}$ \\
\midrule
GPT-5-mini-CoT & 48.0$_{\pm 0.9}$ & \textbf{49.3}$_{\pm 0.4}$ & 54.4$_{\pm 0.3}$ & 64.3$_{\pm 0.2}$ & \textbf{58.3}$_{\pm 7.5}$ & \textbf{79.0}$_{\pm 0.5}$ & 60.4$_{\pm 0.3}$ & 59.0$_{\pm 0.6}$ & \textbf{63.2}$_{\pm 0.4}$ & 83.4$_{\pm 0.7}$ & 56.8$_{\pm 0.9}$ & 61.5$_{\pm 0.8}$ \\
GPT-5-mini-2FT & 43.2$_{\pm 0.4}$ & 45.8$_{\pm 0.5}$ & 51.3$_{\pm 0.6}$ & 58.4$_{\pm 0.6}$ & 54.4$_{\pm 6.4}$ & 74.1$_{\pm 0.7}$ & 56.8$_{\pm 1.1}$ & 58.5$_{\pm 0.6}$ & 55.8$_{\pm 0.1}$ & 84.5$_{\pm 0.2}$ & 56.6$_{\pm 0.4}$ & 58.1$_{\pm 0.5}$ \\
GPT-5-mini-CAT & \textbf{48.7}$_{\pm 0.1}$ & 46.2$_{\pm 0.4}$ & \textbf{55.3}$_{\pm 0.4}$ & \textbf{66.4}$_{\pm 0.3}$ & 55.4$_{\pm 5.6}$ & 76.5$_{\pm 0.5}$ & \textbf{64.5}$_{\pm 11.2}$ & \textbf{63.0}$_{\pm 0.8}$ & 60.6$_{\pm 0.8}$ & \textbf{86.3}$_{\pm 0.5}$ & \textbf{60.9}$_{\pm 0.7}$ & \textbf{62.2}$_{\pm 1.5}$ \\

\bottomrule
\end{tabular}
}

\label{tab:emotion}
\end{adjustwidth}
\end{table}

\section{Results analysis and discussion}\label{sec:results}

\begin{figure*}[!ht]
\begin{adjustwidth}{-2.25in}{0in}
    \centering
    \includegraphics[width=1\linewidth]{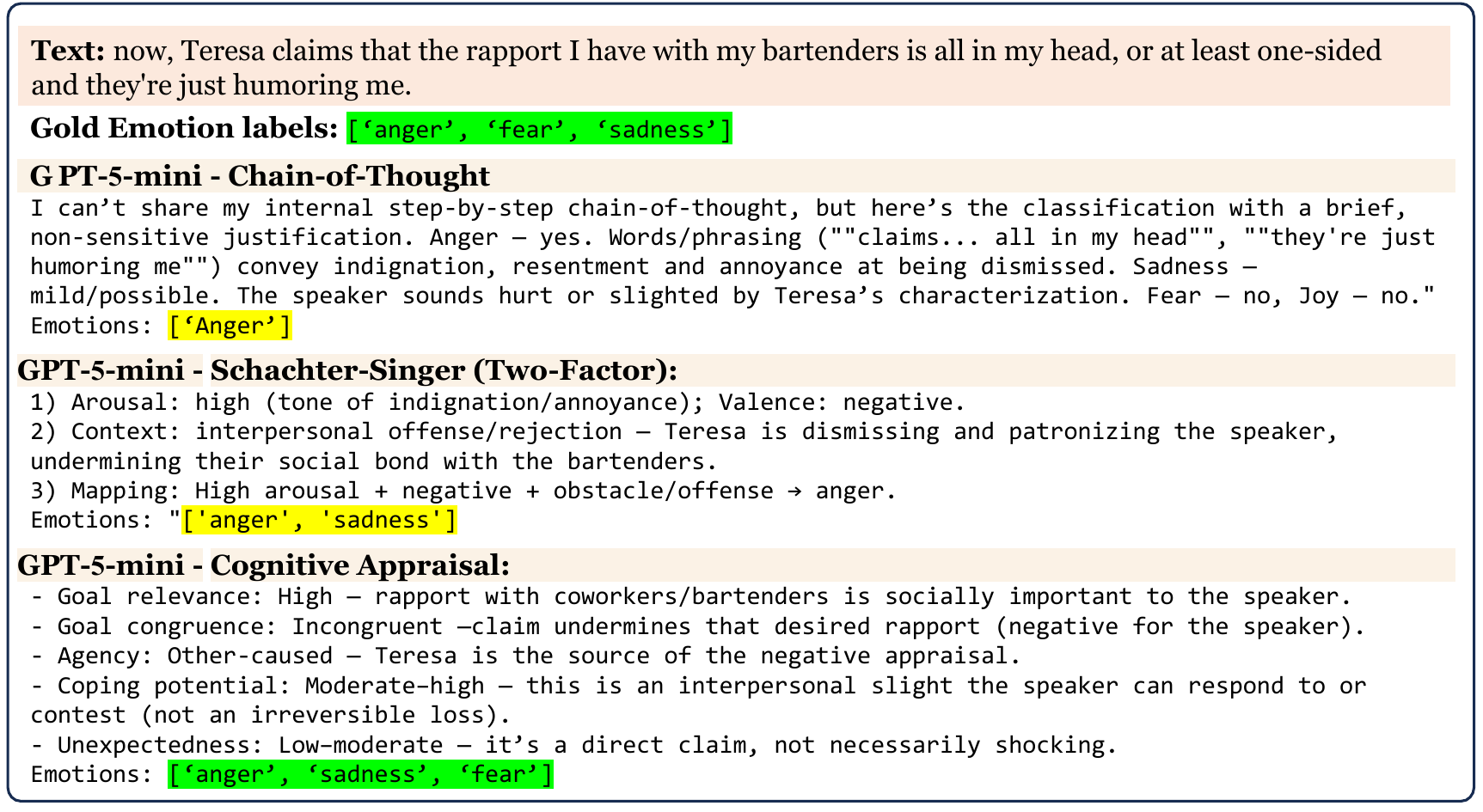}
    \caption{\textbf{Thinking steps output example from the three types of prompts. Chain-of-thought prompting (CoT), the Schachter-Singer (Two-Factor) Theory (2FT), and the Cognitive Appraisal Theory (CAT) prompts, respectively. \texttt{Text} is the English target text asked the LLMs to predict the emotions present in the text, and the output is from GPT-5-mini.}}
    \label{fig:prompt-example}
\end{adjustwidth}
\end{figure*}

\paragraph{Multi-label emotion analysis:} The multi-label emotion analysis results are presented in \Cref{tab:emotion}. As the results show, Cognitive Appraisal Theory-based (CAT) prompting outperforms the baseline CoT and the Schachter-Singer theory-based (2FT) prompt. The superior performance of Cognitive Appraisal-based prompting in multi-label emotion classification can be driven by its ability to resolve emotional ambiguity through contextual evaluation. Unlike general CoT reasoning, which relies on broad semantic logic, Cognitive Appraisal theory provides a structured decision matrix that assesses specific dimensions such as control, agency, and goal conduciveness. This framework enables the model to distinguish between emotions that share similar negative sentiments but differ in their causal structures (e.g., Anger and Fear, based on the presence of a ``control'' thinking step).
By forcing the model to evaluate the ``cognitive evaluation'' of the event rather than just the feeling, cognitive appraisal prompting can effectively create distinct decision boundaries between the overlapping emotion labels. A step-by-step thinking example from the Cognitive Appraisal theory prompt is shown in Fig \ref{fig:prompt-example}.  Among open-weight LLMs, Gemma-3-27B performs better than other evaluated larger LLMs in terms of parameter count, such as Mistral-large-123B and Llama-3-70B. This efficiency may be attributed to Gemma's large (256k-token) vocabulary and greater multilingualism, which enable richer linguistic representation across diverse languages \cite{gemmateam2025gemma3technicalreport}. 

\begin{figure}[!ht]
\begin{adjustwidth}{-2.25in}{0in}
\centering

\begin{subfigure}{0.49\dimexpr\textwidth+2.25in}
\centering
    \includegraphics[width=\linewidth]{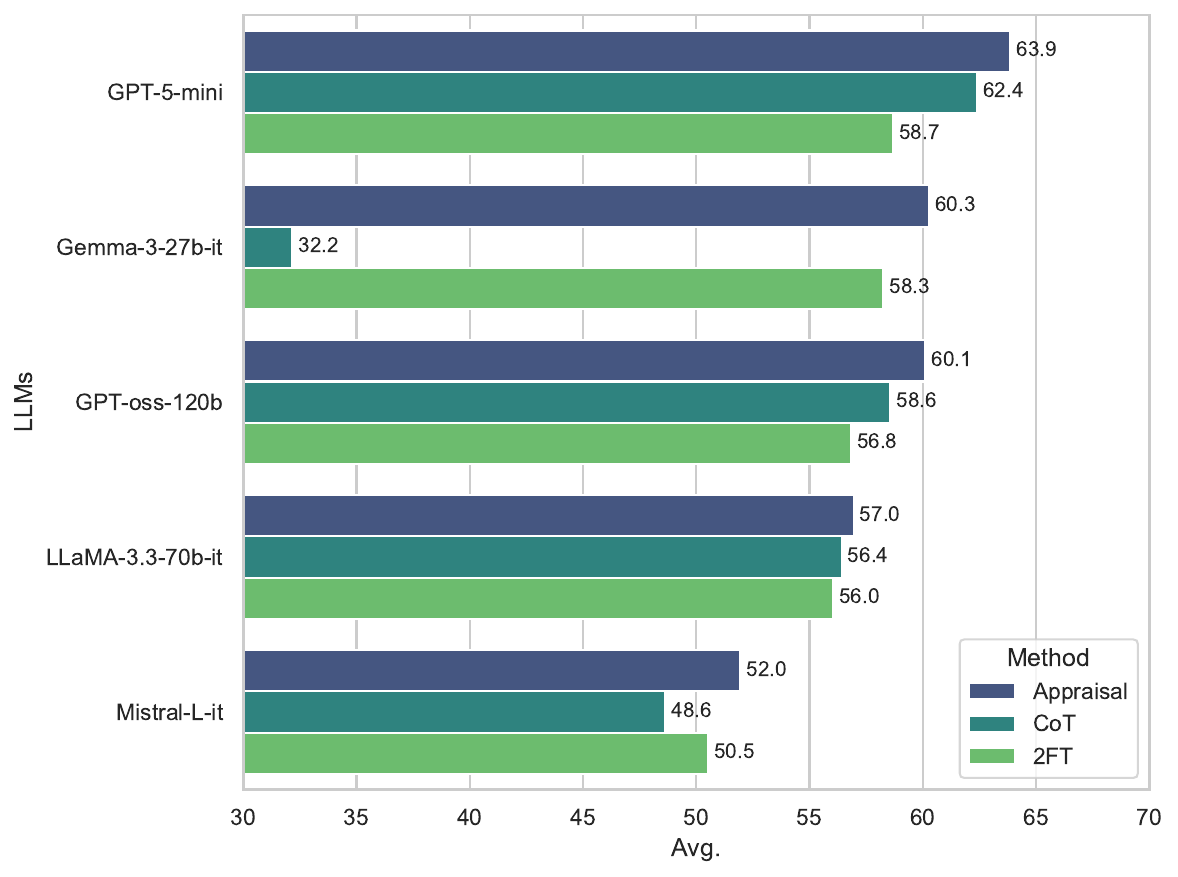}
    \caption{}
    \label{fig:emotion-pilot}
\end{subfigure}
\hfill
\begin{subfigure}{0.49\dimexpr\textwidth+2.25in}
\centering
    \includegraphics[width=\linewidth]{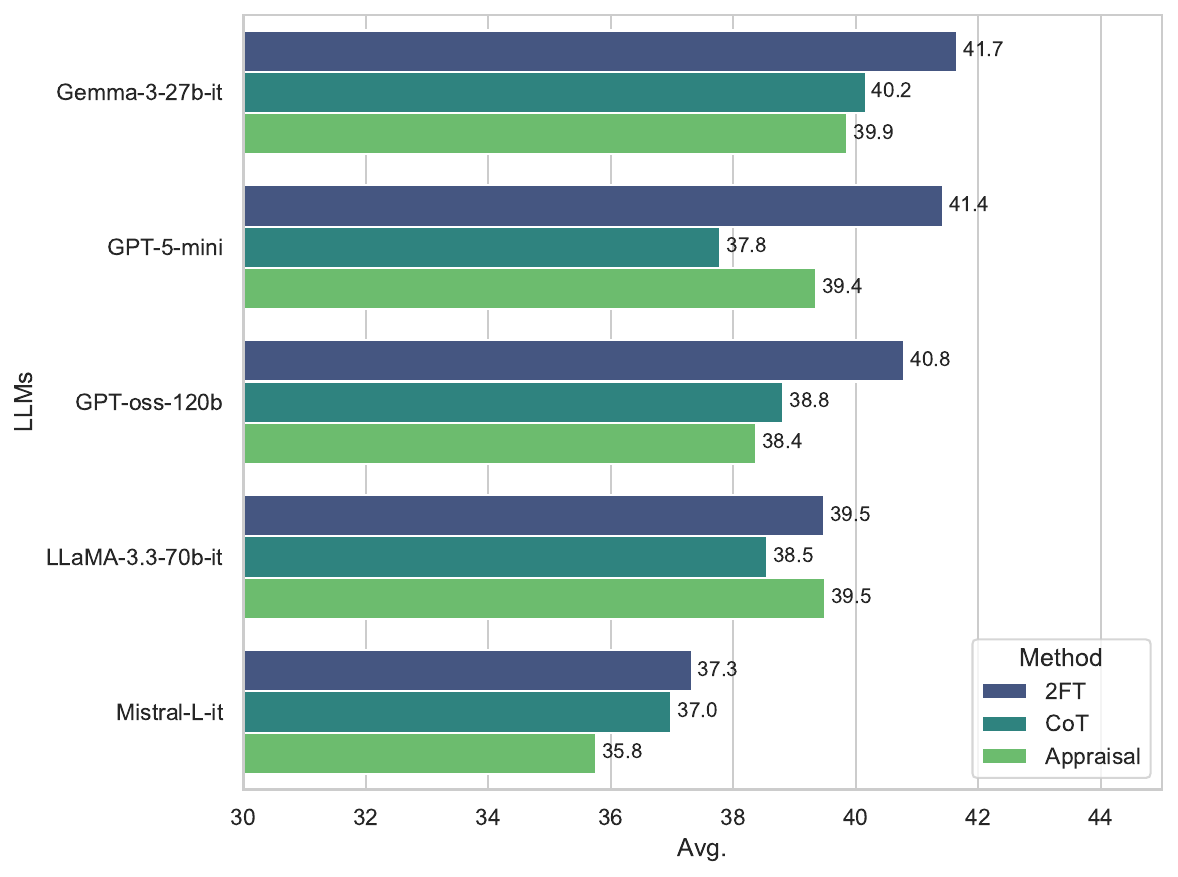}
    \caption{}
    \label{fig:intensity-pilot}
\end{subfigure}

\caption{
\textbf{Average F1 results across languages for emotion and intensity prediction.}
(a) For multi-label emotion analysis, Cognitive Appraisal Theory outperforms prompts based on Chain-of-Thought (CoT) and Schachter-Singer (Two-Factor) theory.
(b) For multi-label emotion intensity analysis, the Schachter-Singer (Two-Factor) theory prompt outperforms Cognitive Appraisal and CoT-based prompts.
}
\label{fig:combined-results}

\end{adjustwidth}
\end{figure}

\paragraph{Emotion intensity prediction} The emotion intensity prediction results are presented in \Cref{tab:intensity}. In the emotion intensity prediction task, the Schachter-Singer theory-based prompt (2FT) outperforms the baseline CoT and the Cognitive Appraisal-based prompt (CAT).  
The superior performance of the Schachter-Singer-based prompt in emotion intensity prediction can be attributed to its explicit isolation of physiological arousal (Factor 1) as a distinct evaluative dimension, effectively decoupling the ``energy" of an emotion from its cognitive context. Standard CoT prompting relies heavily on semantic logic and causality, often conflating the contextual severity of an event (e.g., “failing a test”) with the actual emotional intensity expressed. In contrast, the Schachter–Singer prompt encourages the model to prioritize arousal level (low, medium, and high) and linguistic markers of physiological activation (e.g., lexical modifiers, punctuation, and visceral verbs) before assigning an emotion label. The overall average scores across languages and LLMs for multi-label emotion analysis and intensity prediction are shown in Figs \ref{fig:emotion-pilot} and \ref{fig:intensity-pilot}, respectively. According to the statistical comparison of the standard deviations across three runs, the best-performing model under the 2FT prompting strategy, Gemma-3-27B, exhibits a lower average standard deviation ($\pm 0.2$), indicating greater stability than the \textbf{Chain-of-Thought (CoT) prompting strategy ($\pm 0.4$)}.

\begin{table}[!ht]
\begin{adjustwidth}{-2.25in}{0in}
\centering
\caption{\textbf{Multi-label emotion intensity results using three prompt variants. The baseline chain-of-thought (CoT), the Schachter-Singer Theory (2FT), and the Cognitive Appraisal Theory (CAT) prompts. Results are in zero-shot prompting.}}
\resizebox{\dimexpr\textwidth+2.25in}{!}{
\begin{tabular}{llllllllllll|l}
\midrule
\textbf{Model} & \texttt{amh} & \texttt{arq} & \texttt{chn} & \texttt{deu} & \texttt{eng} & \texttt{esp} & \texttt{hau} & \texttt{ptbr} & \texttt{ron} & \texttt{rus} & \texttt{ukr} & \textbf{Avg.} \\
\midrule
Gemma-3-27B-it-CoT & 32.7$_{\pm 0.8}$ & \textbf{43.2}$_{\pm 0.8}$ & 33.3$_{\pm 0.6}$ & 40.9$_{\pm 0.1}$ & 54.2$_{\pm 1.1}$ & 44.2$_{\pm 0.8}$ & 39.9$_{\pm 0.7}$ & 36.8$_{\pm 0.4}$ & \textbf{41.5}$_{\pm 1.3}$ & 46.6$_{\pm 1.4}$ & 33.1$_{\pm 0.4}$ & 40.6$_{\pm 0.4}$ \\
Gemma-3-27B-it-2FT & \textbf{32.8}$_{\pm 0.7}$ & 41.8$_{\pm 1.5}$ & \textbf{36.7}$_{\pm 0.4}$ & \textbf{42.9}$_{\pm 0.7}$ & \textbf{54.7}$_{\pm 1.9}$ & \textbf{45.4}$_{\pm 0.1}$ & \textbf{37.9}$_{\pm 0.1}$ & \textbf{39.9}$_{\pm 0.8}$ & 41.2$_{\pm 0.3}$ & \textbf{50.6}$_{\pm 0.7}$ & \textbf{36.6}$_{\pm 0.1}$ & \textbf{41.9}$_{\pm 0.2}$ \\
Gemma-3-27B-it-CAT & 30.2$_{\pm 0.4}$ & 41.4$_{\pm 0.4}$ & 32.7$_{\pm 1.6}$ & 40.1$_{\pm 0.4}$ & 49.6$_{\pm 0.2}$ & 42.9$_{\pm 1.0}$ & 37.1$_{\pm 1.3}$ & 35.8$_{\pm 2.8}$ & 38.2$_{\pm 1.0}$ & 44.2$_{\pm 3.1}$ & 32.2$_{\pm 1.1}$ & 38.6$_{\pm 1.1}$ \\
\midrule
LLaMA-3.3-70B-it-CoT & \textbf{25.3}$_{\pm 0.3}$ & 39.6$_{\pm 0.7}$ & 33.7$_{\pm 0.6}$ & 39.7$_{\pm 1.5}$ & 49.6$_{\pm 2.7}$ & 45.3$_{\pm 0.6}$ & 33.1$_{\pm 0.6}$ & 38.2$_{\pm 0.8}$ & 42.9$_{\pm 1.1}$ & 45.9$_{\pm 0.5}$ & 34.7$_{\pm 0.6}$ & 38.9$_{\pm 0.3}$ \\
LLaMA-3.3-70B-it-2FT & 24.8$_{\pm 0.5}$ & 41.0$_{\pm 0.5}$ & \textbf{35.0}$_{\pm 1.3}$ & \textbf{40.3}$_{\pm 1.9}$ & 49.4$_{\pm 1.2}$ & \textbf{46.0}$_{\pm 0.7}$ & 32.0$_{\pm 0.3}$ & \textbf{38.7}$_{\pm 0.6}$ & \textbf{44.3}$_{\pm 0.1}$ & \textbf{47.4}$_{\pm 0.3}$ & \textbf{36.4}$_{\pm 0.2}$ & \textbf{39.6}$_{\pm 0.1}$ \\
LLaMA-3.3-70B-it-CAT & 24.4$_{\pm 0.8}$ & \textbf{41.3}$_{\pm 1.2}$ & 34.1$_{\pm 0.6}$ & 39.6$_{\pm 0.5}$ & \textbf{52.4}$_{\pm 0.7}$ & 45.9$_{\pm 0.1}$ & \textbf{33.1}$_{\pm 0.7}$ & 37.0$_{\pm 0.5}$ & 41.4$_{\pm 0.4}$ & 46.0$_{\pm 0.2}$ & 34.2$_{\pm 1.1}$ & 39.0$_{\pm 0.5}$ \\
\midrule
Mistral-123B-it-CoT & 25.9$_{\pm 1.0}$ & 38.6$_{\pm 0.8}$ & \textbf{31.1}$_{\pm 0.5}$ & 40.0$_{\pm 1.8}$ & 52.6$_{\pm 3.4}$ & 44.2$_{\pm 0.7}$ & \textbf{33.0}$_{\pm 1.8}$ & 36.1$_{\pm 0.5}$ & 38.9$_{\pm 3.4}$ & 42.6$_{\pm 0.6}$ & 32.0$_{\pm 0.9}$ & 37.8$_{\pm 0.7}$ \\
Mistral-123B-it-2FT & \textbf{26.9}$_{\pm 1.0}$ & \textbf{39.6}$_{\pm 0.9}$ & 30.2$_{\pm 2.0}$ & \textbf{42.4}$_{\pm 1.4}$ & \textbf{53.4}$_{\pm 2.6}$ & \textbf{44.9}$_{\pm 0.4}$ & 32.1$_{\pm 2.3}$ & \textbf{36.4}$_{\pm 0.7}$ & \textbf{39.6}$_{\pm 1.4}$ & \textbf{46.0}$_{\pm 0.9}$ & \textbf{32.3}$_{\pm 1.4}$ & \textbf{38.4}$_{\pm 0.6}$ \\
Mistral-123B-it-CAT  & 25.0$_{\pm 1.4}$ & 35.1$_{\pm 1.3}$ & 28.2$_{\pm 0.6}$ & 36.3$_{\pm 0.5}$ & 51.4$_{\pm 1.6}$ & 41.8$_{\pm 0.4}$ & 30.8$_{\pm 2.6}$ & 32.7$_{\pm 0.5}$ & 37.7$_{\pm 0.3}$ & 41.8$_{\pm 1.7}$ & 30.4$_{\pm 1.4}$ & 35.6$_{\pm 0.3}$ \\
\midrule
GPT-oss-120B-CoT & 27.8$_{\pm 0.4}$ & \textbf{42.6}$_{\pm 1.4}$ & 30.7$_{\pm 1.2}$ & 37.5$_{\pm 0.2}$ & \textbf{52.9}$_{\pm 1.2}$ & 45.3$_{\pm 1.1}$ & \textbf{37.2}$_{\pm 0.5}$ & 36.9$_{\pm 1.1}$ & 39.0$_{\pm 0.1}$ & 43.2$_{\pm 0.8}$ & 33.8$_{\pm 1.2}$ & 38.8$_{\pm 0.2}$ \\
GPT-oss-120B-2FT & \textbf{29.8}$_{\pm 0.5}$ & 42.3$_{\pm 0.4}$ & \textbf{32.5}$_{\pm 0.7}$ & \textbf{40.7}$_{\pm 0.4}$ & 51.8$_{\pm 0.6}$ & \textbf{47.3}$_{\pm 0.9}$ & 36.2$_{\pm 0.5}$ & \textbf{38.8}$_{\pm 0.9}$ & \textbf{40.9}$_{\pm 0.1}$ & \textbf{48.0}$_{\pm 1.6}$ & \textbf{38.1}$_{\pm 0.4}$ & \textbf{40.6}$_{\pm 0.3}$ \\
GPT-oss-120B-CAT & 26.8$_{\pm 1.5}$ & 41.5$_{\pm 0.5}$ & 28.2$_{\pm 1.7}$ & 36.8$_{\pm 1.2}$ & 51.3$_{\pm 0.1}$ & 44.5$_{\pm 0.5}$ & 35.4$_{\pm 0.9}$ & 34.6$_{\pm 1.5}$ & 37.4$_{\pm 1.0}$ & 42.8$_{\pm 1.0}$ & 31.8$_{\pm 0.8}$ & 37.4$_{\pm 0.9}$ \\
\midrule
GPT-5-mini-CoT & 30.0$_{\pm 0.2}$ & 43.5$_{\pm 0.8}$ & 28.8$_{\pm 0.5}$ & 36.8$_{\pm 0.3}$ & 54.0$_{\pm 0.3}$ & 42.6$_{\pm 0.3}$ & 38.5$_{\pm 1.0}$ & 33.9$_{\pm 0.4}$ & 39.2$_{\pm 0.2}$ & 40.2$_{\pm 0.7}$ & 30.4$_{\pm 0.2}$ & 38.0$_{\pm 0.3}$ \\
GPT-5-mini-2FT & \textbf{31.7}$_{\pm 1.0}$ & \textbf{44.9}$_{\pm 0.8}$ & \textbf{32.0}$_{\pm 0.3}$ & \textbf{40.8}$_{\pm 0.9}$ & 54.3$_{\pm 0.8}$ & \textbf{47.1}$_{\pm 0.7}$ & \textbf{38.7}$_{\pm 0.6}$ & \textbf{37.5}$_{\pm 1.2}$ & 44.0$_{\pm 0.8}$ & \textbf{47.1}$_{\pm 0.8}$ & \textbf{36.3}$_{\pm 0.4}$ & \textbf{41.3}$_{\pm 0.1}$ \\
GPT-5-mini-CAT & 29.7$_{\pm 0.2}$ & 43.2$_{\pm 0.3}$ & 29.3$_{\pm 0.4}$ & 37.2$_{\pm 0.6}$ & \textbf{54.4}$_{\pm 0.2}$ & 44.8$_{\pm 0.4}$ & 37.3$_{\pm 1.2}$ & 35.4$_{\pm 0.6}$ & \textbf{40.4}$_{\pm 0.6}$ & 44.1$_{\pm 1.3}$ & 32.7$_{\pm 0.6}$ & 39.0$_{\pm 0.3}$ \\

\bottomrule
\end{tabular}
}

\label{tab:intensity}
\end{adjustwidth}
\end{table}

\begin{figure}[!h]
\begin{adjustwidth}{-1.5in}{0in}
    \centering
    \includegraphics[width=1\linewidth]{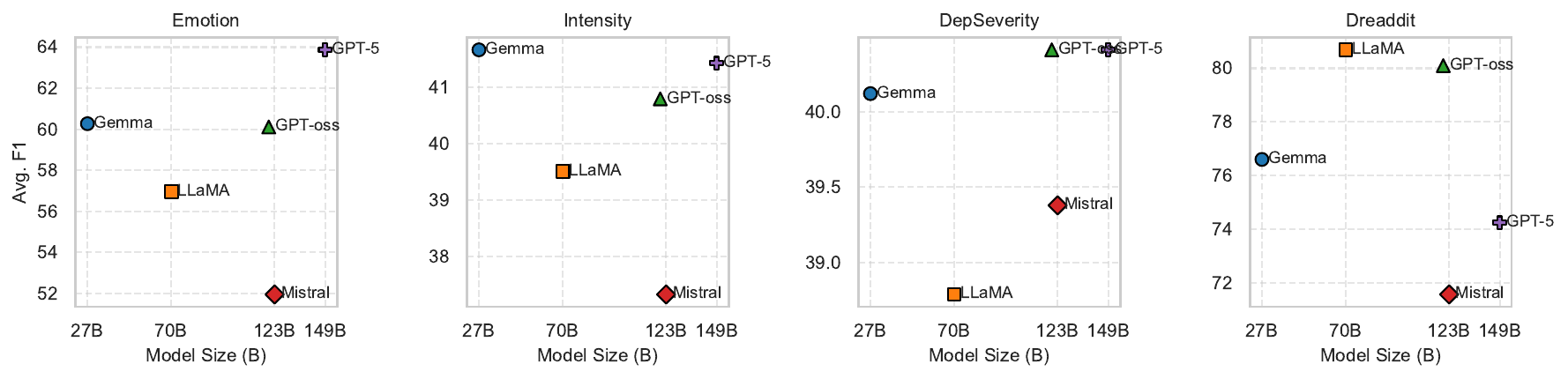}
\caption{\textbf{Average Macro F1 scores across 11 languages for different LLM parameter sizes and evaluation tasks.}}
    \label{fig:best-llm}
\end{adjustwidth}
\end{figure}

\paragraph{Mental health analysis:} The mental health results are presented in \Cref{tab:health}. The proposed PsyCoT outperforms the CoT baseline. For scaling tasks, the Schachter-Singer theory-based Prompt outperforms the baseline CoT and the Cognitive appraisal-based prompt. This is because the Schachter-Singer two-factor theory posits the existence of arousal. For classifications such as stress and multi-label emotion analysis, the Cognitive Appraisal-based prompt outperforms the CoT prompt.   

\begin{table}[!h]
\centering
\caption{\textbf{Stress prediction (Dreaddit) and depression severity (DepSeverity) Macro F1 and accuracy results. Results are in zero-shot prompting.}}

\resizebox{\columnwidth}{!}{
\begin{tabular}{lcc||cc}
\toprule

& \multicolumn{2}{c||}{\textbf{Stress prediction}} 
& \multicolumn{2}{c}{\textbf{Depression severity}} \\

\cmidrule(lr){2-3} \cmidrule(lr){4-5}

\textbf{Model} & \textbf{F1} & \textbf{Acc.} & \textbf{F1} & \textbf{Acc.} \\

\midrule
Gemma-3-27B-it-CoT & 63.1$_{\pm 0.2}$ & 67.2$_{\pm 0.2}$ & 38.1$_{\pm 0.2}$ & 37.3$_{\pm 1.4}$ \\
Gemma-3-27B-it-2FT & \textbf{76.5}$_{\pm 0.1}$ & \textbf{77.3}$_{\pm 0.1}$ & 38.6$_{\pm 0.6}$ & 37.1$_{\pm 0.1}$ \\
Gemma-3-27B-it-CAT & 71.6$_{\pm 2.4}$ & 72.1$_{\pm 0.2}$ & \textbf{39.6}$_{\pm 0.5}$ & \textbf{38.4}$_{\pm 0.6}$ \\

\midrule
LLaMA-3.3-70B-it-CoT & 74.7$_{\pm 0.4}$ & 75.9$_{\pm 0.4}$ & 35.7$_{\pm 0.2}$ & 35.6$_{\pm 0.1}$ \\
LLaMA-3.3-70B-it-2FT & \textbf{80.4}$_{\pm 0.4}$ & \textbf{80.7}$_{\pm 0.4}$ & \textbf{38.2}$_{\pm 0.5}$ & \textbf{37.6}$_{\pm 0.6}$ \\
LLaMA-3.3-70B-it-CAT & 75.0$_{\pm 0.5}$ & 76.0$_{\pm 0.5}$ & 37.9$_{\pm 0.1}$ & 37.2$_{\pm 0.2}$ \\

\midrule
Mistral-123B-it-CoT & 58.3$_{\pm 6.8}$ & 60.9$_{\pm 7.2}$ & 38.8$_{\pm 3.9}$ & 38.5$_{\pm 4.0}$ \\
Mistral-123B-it-2FT & \textbf{72.9}$_{\pm 1.1}$ & \textbf{74.5}$_{\pm 1.2}$ & 38.0$_{\pm 0.3}$ & 37.2$_{\pm 0.3}$ \\
Mistral-123B-it-CAT & 70.6$_{\pm 2.3}$ & 72.8$_{\pm 1.9}$ & \textbf{40.4}$_{\pm 1.5}$ & \textbf{40.0}$_{\pm 1.7}$ \\

\midrule
GPT-oss-120B-CoT & 72.6$_{\pm 0.9}$ & 74.4$_{\pm 0.8}$ & 35.2$_{\pm 0.9}$ & 37.4$_{\pm 0.7}$ \\
GPT-oss-120B-2FT & \textbf{79.5}$_{\pm 0.8}$ & \textbf{79.6}$_{\pm 0.9}$ & 34.8$_{\pm 1.4}$ & 36.6$_{\pm 1.2}$ \\
GPT-oss-120B-CAT & 75.6$_{\pm 0.0}$ & 76.4$_{\pm 0.1}$ & \textbf{40.0}$_{\pm 0.8}$ & \textbf{41.3}$_{\pm 0.7}$ \\

\midrule
GPT-5-mini-CoT & 61.6$_{\pm 0.6}$ & 62.5$_{\pm 0.8}$ & 37.6$_{\pm 0.2}$ & 37.0$_{\pm 0.0}$ \\
GPT-5-mini-2FT & \textbf{74.9}$_{\pm 0.8}$ & \textbf{75.5}$_{\pm 0.9}$ & 37.7$_{\pm 0.9}$ & 37.6$_{\pm 0.4}$ \\
GPT-5-mini-CAT & 71.8$_{\pm 0.8}$ & 73.6$_{\pm 0.6}$ & \textbf{40.0}$_{\pm 0.7}$ & \textbf{39.4}$_{\pm 0.7}$ \\
\bottomrule
\end{tabular}
}
\label{tab:health}
\end{table}

\paragraph{LLMs performance across subjective NLP tasks:} Among the open-weight LLMs, Gemma-3-27B performs better than other open-weight LLMs across the targeted tasks, including models with more parameters such as Llama-3-70B and Mistral-123B. The  GPT-5-mini closed-source LLM outperforms open-weight LLMs. The average score on various LLMs corresponding to their parameter size is shown in Fig \ref{fig:best-llm}.

\paragraph {Low-resource vs. high-resource language results:} Across all evaluated models, performance remains lower. This might be due to the complexity of evaluating subjective tasks.  For low-resource languages such as Amharic (amh) and Hausa (hau), highlighting the persistent challenges LLMs face in underrepresented linguistic settings. In contrast, the open-weight Gemma-3-27B model demonstrates comparatively better performance across both low- and high-resource languages, suggesting better cross-lingual generalization. 

Better performance is achieved for high-resource languages, including English (eng), German (deu), Spanish (spa), Chinese/Mandarin (chn), and Russian (rus), for which training data are substantially more abundant. Smaller models, such as LLaMA-3-8B \cite{grattafiori2024llama3herdmodels}, struggle with detail-oriented, physiology-based prompts, indicating limitations in reasoning depth and instruction-following capacity. Furthermore, the DeepSeek-R1 \cite{DeepSeek-r1} model fails due to output-formatting inconsistencies that undermine evaluation reliability, despite potential gains in reasoning capability.

The emotion-theoretical grounding prompt provides a fine-grained control mechanism that standard logical reasoning lacks, allowing the model to accurately differentiate between high- and low-arousal states even when they share similar semantic polarities \cite{kuhbandner2011dissociable}. As shown in Tables \ref{tab:emotion}, \ref{tab:intensity}, and \ref{tab:health}, the Schachter-Singer and Cognitive Appraisal theory-based prompts (PsyCoT) demonstrate strong efficacy across tasks of varying difficulty and on diverse LLMs, and they outperform existing state-of-the-art CoT prompt engineering approaches. 

\section{Conclusion}\label{sec:conclusion}

In this work, we proposed Psychology-based Chain-of-Thought (PsyCoT), a theory-driven prompting strategy grounded in Schachter–Singer Two-Factor Theory and Cognitive Appraisal Theory of emotion. We evaluated the proposed framework across four downstream subjective NLP tasks: multi-label emotion classification, emotion intensity prediction, stress prediction, and depression severity classification, covering 11 languages. PsyCoT aims to advance emotion understanding in language models through three key capabilities. First, it facilitates the identification of implicit emotions by uncovering emotional states that are not explicitly expressed through emotion-related keywords. Second, it encourages psychologically informed emotion reasoning by guiding models to consider situational antecedents, physiological arousal, and cognitive appraisals before assigning an emotion label. Third, it generates theory-informed textual rationales that explicitly articulate relevant psychological constructs and appraisal factors associated with the predicted emotions.

Experimental results demonstrate that PsyCoT variants offer performance advantages over standard Chain-of-Thought (CoT) prompting, although their effectiveness varies across tasks, models, and languages. In multi-label emotion classification, PsyCoT achieves performance gains of up to 28.1\%, highlighting the benefits of incorporating psychological appraisals into emotion recognition. For multi-label emotion intensity prediction, the Schachter–Singer-based (2FT) variant demonstrates a clearer advantage over both standard CoT and CAT prompting, with improvements of up to 4.8\%. A similar advantage is observed in stress prediction, where 2FT achieves F1-score gains of up to 14.6\% over standard CoT. In contrast, Cognitive Appraisal-based (CAT) prompting is more effective for depression severity classification, achieving the highest F1 scores across most evaluated models. These findings reveal a task-dependent distinction in the contributions of psychological theories to LLM reasoning: cognitive appraisal provides a useful foundation for interpreting contextual and evaluative aspects of emotional experiences, whereas the Schachter–Singer perspective offers advantages in tasks involving emotional intensity and stress-related responses.

Furthermore, our evaluation highlights the importance of model selection in multilingual subjective NLP. In particular, Gemma-3-27B demonstrates competitive performance compared with substantially larger models, including Llama-3-70B and Mistral-123B, suggesting that model size alone does not determine effectiveness in psychologically grounded reasoning tasks.

Overall, by bridging established psychological theories with computational linguistics, PsyCoT moves beyond surface-level emotion classification toward more theory-informed and interpretable emotion reasoning. Our findings suggest that incorporating complementary psychological perspectives into prompting strategies can improve the performance of LLMs across diverse emotion and mental-health NLP tasks, while providing a foundation for more transparent and psychologically informed affective computing systems.

\section*{Limitations}

This work is not without limitations. Although the proposed psychology-grounded reasoning approach (PsyCoT) achieves promising results, several limitations should be acknowledged. First, our evaluation primarily relies on automatic performance metrics, such as F1 score and accuracy, and does not include formal statistical significance testing. Future work should incorporate appropriate significance tests to determine whether the observed improvements are statistically reliable.

Second, PsyCoT is currently grounded in two well-established theories of emotion: the Schachter-Singer theory and Cognitive Appraisal theory. Although these theories provide complementary perspectives on emotional processes, they do not capture the full range of psychological theories of emotion. Future research could investigate whether incorporating additional emotion theories leads to further improvements in subjective NLP tasks.

Third, we evaluate PsyCoT across four subjective NLP tasks and a limited set of datasets. Extending the evaluation to a broader range of subjective NLP tasks, larger and more diverse datasets, and additional languages would provide a more comprehensive assessment of the approach. In particular, future studies should examine PsyCoT's effectiveness with larger sample sizes to assess its explainability and generalizability.

Fourth, although PsyCoT generates psychologically structured explanations, an important distinction should be made between theory-informed explanation generation and validated explainability. By explicitly incorporating psychological constructs from Schachter–Singer Two-Factor Theory and Cognitive Appraisal Theory, PsyCoT encourages LLMs to articulate relevant contextual and appraisal-related factors underlying their predictions. This provides a more structured and theoretically grounded account of the factors considered during prediction. However, psychological experts have not independently evaluated the generated rationales to establish their psychological validity, nor have we conducted faithfulness analyses to determine whether they accurately reflect the models' underlying decision-making processes. Consequently, these explanations should be interpreted as theory-guided textual rationales rather than verified accounts of human-like psychological reasoning or faithful representations of internal model processes.

Finally, our experiments primarily use English-language prompts, including when evaluating non-English languages. This may introduce a potential source of linguistic and cultural bias. Future work should investigate language-specific prompting, where both the prompts and the reasoning instructions are provided in the target language. For example, when evaluating Amharic, the prompting and reasoning instructions could be formulated in Amharic rather than English. Such multilingual prompting could provide a more linguistically and culturally appropriate evaluation of subjective language understanding.






\bibliography{plos_bibtex_sample}

@article{bty842,
    author = {Hu, Hailin and Xiao, An and Zhang, Sai and Li, Yangyang and Shi, Xuanling and Jiang, Tao and Zhang, Linqi and Zhang, Lei and Zeng, Jianyang},
    title = {DeepHINT: understanding HIV-1 integration via deep learning with attention},
    journal = {Bioinformatics},
    volume = {35},
    number = {10},
    pages = {1660-1667},
    year = {2019},
    month = {05},
    issn = {1367-4803},
    url = {https://doi.org/10.1093/bioinformatics/bty842},
}

@inproceedings{miglani-etal-2023-using,
    title = "Using Captum to Explain Generative Language Models",
    author = "Miglani, Vivek  and
      Yang, Aobo  and
      Markosyan, Aram  and
      Garcia-Olano, Diego  and
      Kokhlikyan, Narine",
    editor = "Tan, Liling  and
      Milajevs, Dmitrijs  and
      Chauhan, Geeticka  and
      Gwinnup, Jeremy  and
      Rippeth, Elijah",
    booktitle = "Proceedings of the 3rd Workshop for Natural Language Processing Open Source Software (NLP-OSS 2023)",
    month = dec,
    year = "2023",
    address = "Singapore",
    publisher = "Association for Computational Linguistics",
    url = "https://aclanthology.org/2023.nlposs-1.19/",
    pages = "165--173",
}

@inproceedings{depseverty,
author = {Naseem, Usman and Dunn, Adam G. and Kim, Jinman and Khushi, Matloob},
title = {{Early Identification of Depression Severity Levels on Reddit Using Ordinal Classification}},
year = {2022},
isbn = {9781450390965},
publisher = {Association for Computing Machinery},
address = {New York, NY, USA},
url = {https://doi.org/10.1145/3485447.3512128},
booktitle = {Proceedings of the ACM Web Conference 2022},
pages = {2563–2572},
numpages = {10},
location = {Virtual Event, Lyon, France},
series = {WWW '22}
}

@article{SMITH2020107925,
title = {{The evolution and development of the uniquely human capacity for emotional awareness: A synthesis of comparative anatomical, cognitive, neurocomputational, and evolutionary psychological perspectives}},
journal = {Biological Psychology},
volume = {154},
pages = {107925},
year = {2020},
issn = {0301-0511},
doi = {https://doi.org/10.1016/j.biopsycho.2020.107925},
author = {Ryan Smith and Horst Dieter Steklis and Netzin G. Steklis and Karen L. Weihs and Richard D. Lane},
}

@inproceedings{enomoto-etal-2025-fair,
    title = {{A Fair Comparison without Translationese: {E}nglish vs. Target-language Instructions for Multilingual {LLM}s}},
    author = "Enomoto, Taisei  and
      Kim, Hwichan  and
      Chen, Zhousi  and
      Komachi, Mamoru",
    booktitle = "Proceedings of the 2025 Conference of the Nations of the Americas Chapter of the Association for Computational Linguistics: Human Language Technologies (Volume 2: Short Papers)",
    month = apr,
    year = "2025",
    address = "Albuquerque, New Mexico",
    publisher = "Association for Computational Linguistics",
    url = "https://aclanthology.org/2025.naacl-short.55/",
    pages = "649--670",
    ISBN = "979-8-89176-190-2",
}

@misc{losttranslationlargelanguage,
      title={{Lost in Translation: Large Language Models in Non-English Content Analysis}}, 
      author={Gabriel Nicholas and Aliya Bhatia},
      year={2023},
      url={https://arxiv.org/abs/2306.07377}, 
}

@misc{chainofempathyrl,
      title={{Empathy-R1: A Chain-of-Empathy and Reinforcement Learning Framework for Long-Form Mental Health Support}}, 
      author={Xianrong Yao and Dong She and Chenxu Zhang and Yimeng Zhang and Yueru Sun and Noman Ahmed and Yang Gao and Zhanpeng Jin},
      year={2025},
      url={https://arxiv.org/abs/2509.14851}, 
}

@article{chain-of-emphathy,
  author    = {Yoon-Kyung Lee and Sowon Hahn and Seoyeon Bae and Inju Lee and Minjung Shin},
  title     = {{Enhancing Empathic Reasoning of Large Language Models Based on Psychotherapy Models for AI-assisted Social Support}},
  journal   = {Korean Journal of Cognitive Science},
  volume    = {35},
doi={10.19066/cogsci.2024.35.1.002},
  number    = {1},
  pages     = {23-48},
  year      = 2024,
  month     = 03
}

@book{ortony2022cognitive,
  title={{The cognitive structure of emotions}},
  author={Ortony, Andrew and Clore, Gerald L and Collins, Allan},
  year={2022},
url = {https://doi.org/10.1017/CBO9780511571299},
  publisher={Cambridge university press}
}

@inproceedings{yeo-jaidka-2023-peace,
    title = {{The {PEACE}-Reviews dataset: Modeling Cognitive Appraisals in Emotion Text Analysis}},
    author = "Yeo, Gerard  and
      Jaidka, Kokil",
    booktitle = "Findings of the Association for Computational Linguistics: EMNLP 2023",
    month = dec,
    year = "2023",
    address = "Singapore",
    publisher = "Association for Computational Linguistics",
    url = "https://aclanthology.org/2023.findings-emnlp.186/",
    pages = "2822--2840",
}

@inproceedings{yeo-jaidka-2025-beyond,
    title = {{Beyond Context to Cognitive Appraisal: Emotion Reasoning as a Theory of Mind Benchmark for Large Language Models}},
    author = "Yeo, Gerard Christopher  and
      Jaidka, Kokil",
    booktitle = "Findings of the Association for Computational Linguistics: ACL 2025",
    month = jul,
    year = "2025",
    address = "Vienna, Austria",
    publisher = "Association for Computational Linguistics",
    url = "https://aclanthology.org/2025.findings-acl.1359/",
    pages = "26517--26525",
    ISBN = "979-8-89176-256-5",
}

@inproceedings{turcan-mckeown-2019-dreaddit,
    title = {{{D}readdit: A {R}eddit Dataset for Stress Analysis in Social Media}},
    author = "Turcan, Elsbeth  and
      McKeown, Kathy",
    booktitle = "Proceedings of the Tenth International Workshop on Health Text Mining and Information Analysis (LOUHI 2019)",
    month = nov,
    year = "2019",
    address = "Hong Kong",
    publisher = "Association for Computational Linguistics",
    url = "https://aclanthology.org/D19-6213/",
    pages = "97--107",
}

@inproceedings{zhan-etal-2023-evaluating,
    title = {{Evaluating Subjective Cognitive Appraisals of Emotions from Large Language Models}},
    author = "Zhan, Hongli  and
      Ong, Desmond C.  and
      Li, Junyi Jessy",
    booktitle = "Findings of the Association for Computational Linguistics: EMNLP 2023",
    month = dec,
    year = "2023",
    address = "Singapore",
    publisher = "Association for Computational Linguistics",
    url = "https://aclanthology.org/2023.findings-emnlp.962",
    pages = "14418--14446",
}

@inproceedings{mohammad-etal-2018-semeval,
    title = {{{S}em{E}val-2018 Task 1: Affect in Tweets}},
    author = "Mohammad, Saif  and
      Bravo-Marquez, Felipe  and
      Salameh, Mohammad  and
      Kiritchenko, Svetlana",
    booktitle = "Proceedings of the 12th International Workshop on Semantic Evaluation",
    month = jun,
    year = "2018",
    address = "New Orleans, Louisiana",
    publisher = "Association for Computational Linguistics",
    url = "https://aclanthology.org/S18-1001/",
    pages = "1--17",
}

@inproceedings{chae-etal-2023-dialogue,
    title = {{Dialogue Chain-of-Thought Distillation for Commonsense-aware Conversational Agents}},
    author = "Chae, Hyungjoo  and
      Song, Yongho  and
      Ong, Kai  and
      Kwon, Taeyoon  and
      Kim, Minjin  and
      Yu, Youngjae  and
      Lee, Dongha  and
      Kang, Dongyeop  and
      Yeo, Jinyoung",
    booktitle = "Proceedings of the 2023 Conference on Empirical Methods in Natural Language Processing",
    month = dec,
    year = "2023",
    address = "Singapore",
    publisher = "Association for Computational Linguistics",
    url = "https://aclanthology.org/2023.emnlp-main.342/",
    pages = "5606--5632",
}

@inproceedings{NEURIPS2022_18abbeef,
 author = {Lewkowycz, Aitor and Andreassen, Anders and Dohan, David and Dyer, Ethan and Michalewski, Henryk and Ramasesh, Vinay and Slone, Ambrose and Anil, Cem and Schlag, Imanol and Gutman-Solo, Theo and Wu, Yuhuai and Neyshabur, Behnam and Gur-Ari, Guy and Misra, Vedant},
 booktitle = {Advances in Neural Information Processing Systems},
 pages = {3843--3857},
 publisher = {Curran Associates, Inc.},
 title = {{Solving Quantitative Reasoning Problems with Language Models}},
 url = {https://dl.acm.org/doi/10.5555/3600270.3600548},
 volume = {35},
 year = {2022}
}

@misc{Barez-undated-xk,
  title={{Chain-of-thought is not explainability}},
  author       = "Barez, Fazl and Wu, Yu and Lan, Michael and Wang, Vincent and Google, Noah Siegel and Ntu, Clement Neo and Paren, Alasdair
                  and Bibi, Adel and Trager, Robert and Fornasiere, Damiano and Yan, John and Elazar, Yanai and Bengio, Yoshua",
url={https://arxiv.org/abs/2512.23032},
  pages={v1},
  year={2026}
}

@article{vsimic2021understanding,
  title={{Understanding emotions: origins and roles of the amygdala}},
  author={{\v{S}}imi{\'c}, Goran and Tkal{\v{c}}i{\'c}, Mladenka and Vuki{\'c}, Vana and Mulc, Damir and {\v{S}}pani{\'c}, Ena and {\v{S}}agud, Marina and Olucha-Bordonau, Francisco E and Vuk{\v{s}}i{\'c}, Mario and R. Hof, Patrick},
  journal={Biomolecules},
  volume={11},
doi={10.3390/biom11060823},
  number={6},
  pages={823},
  year={2021},
  publisher={MDPI}
}

@article{crivelli2024emotional,
  title={{Emotional and cognitive “route” in decision-making process: The relationship between executive functions, psychophysiological correlates, decisional styles, and personality}},
  author={Crivelli, Davide and Acconito, Carlotta and Balconi, Michela},
  journal={Brain Sciences},
  volume={14},
url={https://doi.org/10.3390/brainsci14070734},
  number={7},
  pages={734},
  year={2024},
  publisher={MDPI}
}

@article{holmes2025applications,
  title={{Applications of large language models in the field of suicide prevention: Scoping review}},
  author={Holmes, Glenn and Tang, Biya and Gupta, Sunil and Venkatesh, Svetha and Christensen, Helen and Whitton, Alexis},
  journal={Journal of Medical Internet Research},
  volume={27},
  pages={e63126},
url={https://www.jmir.org/2025/1/e63126},
  year={2025},
  publisher={JMIR Publications Toronto, Canada}
}

@article{SETTANNI2025116583,
title = {{Assessing the accuracy and consistency of large language models in triaging social media posts for psychological distress}},
journal = {Psychiatry Research},
volume = {351},
pages = {116583},
year = {2025},
issn = {0165-1781},
doi = {https://doi.org/10.1016/j.psychres.2025.116583},
author = {Michele Settanni and Francesco Quilghini and Anna Toscano and Davide Marengo},
}

@article{weber2025using,
  title={{Using a fine-tuned large language model for symptom-based depression evaluation}},
  author={Weber, Samantha and Deperrois, Nicolas and Heun, Robert and Fr{\"u}hsch{\"u}tz, Laura and Monn, Anna and Homan, Stephanie and H{\"a}fliger, Andrea and Seifritz, Erich and Kowatsch, Tobias and MULTICAST consortium J{\"a}ger Lena 8 Schultebraucks Katharina 9 Gershov Sapir 9 Mocellin Jacopo 1 4 and others},
  journal={npj Digital Medicine},
  volume={8},
  number={1},
url={https://www.nature.com/articles/s41746-025-01982-8},
  pages={598},
  year={2025},
  publisher={Nature Publishing Group UK London}
}

@article{sharma2025assessment,
  title={{Assessment of Association of Mental Health to Emotional Intelligence in Medical Undergraduate Students Based On Gender Differences}},
  author={Sharma, Meenakshi and Sharma, Sushil Kumar},
  journal={European Journal of Cardiovascular Medicine},
  volume={15},
  pages={97--102},
url={10.61336/ejcm/25-12-18},
  year={2025}
}

@misc{zhang2025affectivecomputingeralarge,
      title={{Affective Computing in the Era of Large Language Models: A Survey from the NLP Perspective}}, 
      author={Yiqun Zhang and Xiaocui Yang and Xingle Xu and Zeran Gao and Yijie Huang and Shiyi Mu and Shi Feng and Daling Wang and Yifei Zhang and Kaisong Song and Ge Yu},
      year={2025},
      primaryClass={cs.CL},
      url={https://arxiv.org/abs/2408.04638}, 
}

@misc{chen2024recentadvancement,
      title={{Recent Advancement of Emotion Cognition in Large Language Models}}, 
      author={Yuyan Chen and Yanghua Xiao},
      year={2024},
      primaryClass={cs.CL},
      url={https://arxiv.org/abs/2409.13354}, 
}

@misc{song2025largelanguagemodelssubjective,
      title={{Large Language Models for Subjective Language Understanding: A Survey}}, 
      author={Changhao Song and Yazhou Zhang and Hui Gao and Ben Yao and Peng Zhang},
      year={2025},
      primaryClass={cs.CL},
      url={https://arxiv.org/abs/2508.07959}, 
}

@article{DONG2025101014,
title = {{Integrating large language models and affective computing for human-machine symbiosis in intelligent driving}},
journal = {The Innovation},
volume = {6},
number = {12},
pages = {101014},
year = {2025},
issn = {2666-6758},
doi = {https://doi.org/10.1016/j.xinn.2025.101014},
author = {Zhiwu Dong and Chuqiao Chen and Chenlei Liao and Xiqun (Michael) Chen}
}

@INPROCEEDINGS{10970300,
  author={Bhattacharyya, Sree and Yang, Shuhua and Wang, James Z.},
  booktitle={2024 12th International Conference on Affective Computing and Intelligent Interaction (ACII)}, 
  title={{A Heterogeneous Multimodal Graph Learning Framework for Recognizing User Emotions in Social Networks}}, 
  year={2024},
  volume={},
  number={},
  pages={327-336},
  doi={10.1109/ACII63134.2024.00042}
}

@article{yeo2024associations,
  title={{Associations between cognitive appraisals and emotions: A meta-analytic review.}},
  author={Yeo, Gerard C and Ong, Desmond C},
  journal={Psychological Bulletin},
  year={2024},
doi={https://doi.org/10.1037/bul0000452},
  publisher={American Psychological Association}
}

@misc{vatsal2024surveypromptengineeringmethods,
      title={{A Survey of Prompt Engineering Methods in Large Language Models for Different NLP Tasks}}, 
      author={Shubham Vatsal and Harsh Dubey},
      year={2024},
      archivePrefix={arXiv},
      primaryClass={cs.CL},
      url={https://arxiv.org/abs/2407.12994}, 
}

@InProceedings{ramnath-etal-2025-systematic,
title = {{A Systematic Survey of Automatic Prompt Optimization Techniques}},
author = "Ramnath, Kiran and
Zhou, Kang and
Guan, Sheng and
Mishra, Soumya Smruti and
Qi, Xuan and
Shen, Zhengyuan and
Wang, Shuai and
Woo, Sangmin and
Jeoung, Sullam and
Wang, Yawei and
Wang, Haozhu and
Ding, Han and
Lu, Yuzhe and
Xu, Zhichao and
Zhou, Yun and
Srinivasan, Balasubramaniam and
Yan, Qiaojing and
Chen, Yueyan and
Ding, Haibo and
Xu, Panpan and
Cheong, Lin Lee",
booktitle = "Proceedings of the 2025 Conference on Empirical Methods in Natural Language Processing",
month = nov,
year = "2025",
address = "Suzhou, China",
publisher = "Association for Computational Linguistics",
pages = "33066--33098",
url = "https://aclanthology.org/2025.emnlp-main.1681",
}

@article{Glasser1964,
author = {William Glasser},
title ={{Reality Therapy: A Realistic Approach to the Young Offender}},
journal = {Crime \& Delinquency},
volume = {10},
number = {2},
pages = {135-144},
year = {1964},
URL = { 
        https://doi.org/10.1177/001112876401000206
},
}

@article{kuhbandner2011dissociable,
  title={{Dissociable effects of valence and arousal in adaptive executive control}},
  author={Kuhbandner, Christof and Zehetleitner, Michael},
  journal={PloS one},
  volume={6},
  number={12},
  pages={e29287},
  year={2011},
  url={https://doi.org/10.1371/journal.pone.0029287},
  publisher={Public Library of Science San Francisco, USA}
}

@article{cooper2011person,
  title={{Person-centered therapy: A pluralistic perspective}},
  author={Cooper, Mick and McLeod, John},
  journal={Person-Centered \& Experiential Psychotherapies},
  volume={10},
  number={3},
  pages={210--223},
  year={2011},
  url={https://doi.org/10.1080/14779757.2011.599517},
  publisher={Taylor \& Francis}
}

@article{linehan1987dialectical,
  title={{Dialectical behavioral therapy: A cognitive behavioral approach to parasuicide}},
  author={Linehan, Marsha M},
  journal={Journal of Personality disorders},
  volume={1},
  number={4},
  url={https://doi.org/10.1521/pedi.1987.1.4.328},
  pages={328--333},
  year={1987},
  publisher={Guilford Press}
}

@book{beck1979cognitive,
  author    = {Beck, Aaron T.},
  title     = {Cognitive Therapy and the Emotional Disorders},
  year      = {1976},
  publisher = {International Universities Press},
  address   = {New York},
  url={https://openlibrary.org/books/OL10275388M/}
}

@article{maity2025large,
  title={{Large Language Models in Healthcare and Medical Applications: A Review}},
  author={Maity, Subhankar and Saikia, Manob Jyoti},
  journal={Bioengineering},
  volume={12},
url={https://doi.org/10.3390/bioengineering12060631},
  number={6},
  pages={631},
  year={2025},
  publisher={MDPI}
}

@inproceedings{emollms,
author = {Liu, Zhiwei and Yang, Kailai and Xie, Qianqian and Zhang, Tianlin and Ananiadou, Sophia},
title = {{EmoLLMs: A Series of Emotional Large Language Models and Annotation Tools for Comprehensive Affective Analysis}},
year = {2024},
isbn = {9798400704901},
publisher = {Association for Computing Machinery},
address = {New York, NY, USA},
url = {https://doi.org/10.1145/3637528.3671552},
booktitle = {Proceedings of the 30th ACM SIGKDD Conference on Knowledge Discovery and Data Mining},
pages = {5487–5496},
numpages = {10},
location = {Barcelona, Spain},
series = {KDD '24}
}

@article{lazarus1991progress,
  title={{Progress on a cognitive-motivational-relational theory of emotion.}},
  author={Lazarus, Richard S},
  journal={American psychologist},
  volume={46},
  number={8},
url={10.1037//0003-066x.46.8.819},
  pages={819},
  year={1991},
  publisher={American Psychological Association}
}

@article{schachter1962cognitive,
  title={{Cognitive, social, and physiological determinants of emotional state.}},
  author={Schachter, Stanley and Singer, Jerome},
  journal={Psychological review},
  volume={69},
  number={5},
url={https://doi.org/10.1037/h0046234},
  pages={379},
  year={1962},
  publisher={American Psychological Association}
}

@inproceedings{chu-etal-2025-llm,
    title = {{{LLM} Agents for Education: Advances and Applications}},
    author = "Chu, Zhendong  and
      Wang, Shen  and
      Xie, Jian  and
      Zhu, Tinghui  and
      Yan, Yibo  and
      Ye, Jingheng  and
      Zhong, Aoxiao  and
      Hu, Xuming  and
      Liang, Jing  and
      Yu, Philip S.  and
      Wen, Qingsong",
    booktitle = "Findings of the Association for Computational Linguistics: EMNLP 2025",
    month = nov,
    year = "2025",
    address = "Suzhou, China",
    publisher = "Association for Computational Linguistics",
    url = "https://aclanthology.org/2025.findings-emnlp.743/",
    pages = "13782--13810",
    ISBN = "979-8-89176-335-7",
}

@inproceedings{belay-etal-2025-evaluating,
    title = {{Evaluating the Capabilities of Large Language Models for Multi-label Emotion Understanding}},
    author = "Belay, Tadesse Destaw  and
      Azime, Israel Abebe  and
      Ayele, Abinew Ali  and
      Sidorov, Grigori  and
      Klakow, Dietrich  and
      Slusallek, Philip  and
      Kolesnikova, Olga  and
      Yimam, Seid Muhie",
    booktitle = "Proceedings of the 31st International Conference on Computational Linguistics",
    month = jan,
    year = "2025",
    address = "Abu Dhabi, UAE",
    publisher = "Association for Computational Linguistics",
    url = "https://aclanthology.org/2025.coling-main.237/",
    pages = "3523--3540",
}

@inproceedings{fewshot-learners,
author = {Brown, Tom B. and Mann, Benjamin and Ryder, Nick and Subbiah, Melanie and Kaplan, Jared and Dhariwal, Prafulla and Neelakantan, Arvind and Shyam, Pranav and Sastry, Girish and Askell, Amanda and Agarwal, Sandhini and Herbert-Voss, Ariel and Krueger, Gretchen and Henighan, Tom and Child, Rewon and Ramesh, Aditya and Ziegler, Daniel M. and Wu, Jeffrey and Winter, Clemens and Hesse, Christopher and Chen, Mark and Sigler, Eric and Litwin, Mateusz and Gray, Scott and Chess, Benjamin and Clark, Jack and Berner, Christopher and McCandlish, Sam and Radford, Alec and Sutskever, Ilya and Amodei, Dario},
title = {{Language models are few-shot learners}},
year = {2020},
isbn = {9781713829546},
url={https://dl.acm.org/doi/abs/10.5555/3495724.3495883},
publisher = {Curran Associates Inc.},
address = {Red Hook, NY, USA},
booktitle = {Proceedings of the 34th International Conference on Neural Information Processing Systems},
articleno = {159},
numpages = {25},
location = {Vancouver, BC, Canada},
series = {NIPS '20}
}

@article{PRATAP2025100144,
title = {{The fine art of fine-tuning: A structured review of advanced LLM fine-tuning techniques}},
journal = {Natural Language Processing Journal},
volume = {11},
pages = {100144},
year = {2025},
issn = {2949-7191},
doi = {https://doi.org/10.1016/j.nlp.2025.100144},
author = {Samar Pratap and Alston Richard Aranha and Divyanshu Kumar and Gautam Malhotra and Anantharaman Palacode Narayana Iyer and Shylaja S.S.},
}

@inproceedings{muhammad-etal-2025-brighter,
    title = "{BRIGHTER}: {BRI}dging the Gap in Human-Annotated Textual Emotion Recognition Datasets for 28 Languages",
    author = "Muhammad, Shamsuddeen Hassan  and
      Ousidhoum, Nedjma  and
      Abdulmumin, Idris  and
      Wahle, Jan Philip  and
      Ruas, Terry  and
      Beloucif, Meriem  and
      de Kock, Christine  and
      Surange, Nirmal  and
      Teodorescu, Daniela  and
      Ahmad, Ibrahim Said  and
      Adelani, David Ifeoluwa  and
      Aji, Alham Fikri  and
      Ali, Felermino D. M. A.  and
      Alimova, Ilseyar  and
      Araujo, Vladimir  and
      Babakov, Nikolay  and
      Baes, Naomi  and
      Bucur, Ana-Maria  and
      Bukula, Andiswa  and
      Cao, Guanqun  and
      Tufi{\~n}o, Rodrigo  and
      Chevi, Rendi  and
      Chukwuneke, Chiamaka Ijeoma  and
      Ciobotaru, Alexandra  and
      Dementieva, Daryna  and
      Gadanya, Murja Sani  and
      Geislinger, Robert  and
      Gipp, Bela  and
      Hourrane, Oumaima  and
      Ignat, Oana  and
      Lawan, Falalu Ibrahim  and
      Mabuya, Rooweither  and
      Mahendra, Rahmad  and
      Marivate, Vukosi  and
      Panchenko, Alexander  and
      Piper, Andrew  and
      Ferreira, Charles Henrique Porto  and
      Protasov, Vitaly  and
      Rutunda, Samuel  and
      Shrivastava, Manish  and
      Udrea, Aura Cristina  and
      Wanzare, Lilian Diana Awuor  and
      Wu, Sophie  and
      Wunderlich, Florian Valentin  and
      Zhafran, Hanif Muhammad  and
      Zhang, Tianhui  and
      Zhou, Yi  and
      Mohammad, Saif M.",
    editor = "Che, Wanxiang  and
      Nabende, Joyce  and
      Shutova, Ekaterina  and
      Pilehvar, Mohammad Taher",
    booktitle = "Proceedings of the 63rd Annual Meeting of the Association for Computational Linguistics (Volume 1: Long Papers)",
    month = jul,
    year = "2025",
    address = "Vienna, Austria",
    publisher = "Association for Computational Linguistics",
    url = "https://aclanthology.org/2025.acl-long.436/",
    pages = "8895--8916",
    ISBN = "979-8-89176-251-0",
}

@inproceedings{dong-etal-2024-survey,
    title = {{A Survey on In-context Learning}},
    author = "Dong, Qingxiu  and
      Li, Lei  and
      Dai, Damai  and
      Zheng, Ce  and
      Ma, Jingyuan  and
      Li, Rui  and
      Xia, Heming  and
      Xu, Jingjing  and
      Wu, Zhiyong  and
      Chang, Baobao  and
      Sun, Xu  and
      Li, Lei  and
      Sui, Zhifang",
    booktitle = "Proceedings of the 2024 Conference on Empirical Methods in Natural Language Processing",
    month = nov,
    year = "2024",
    address = "Miami, Florida, USA",
    publisher = "Association for Computational Linguistics",
    url = "https://aclanthology.org/2024.emnlp-main.64/",
    doi = "10.18653/v1/2024.emnlp-main.64",
    pages = "1107--1128",
}

@article{bai2025holistic,
  title={{A holistic comparative study of large language models as emotional support dialogue systems}},
  author={Bai, Xin and Chen, Guanyi and He, Tingting and Zhou, Chenlian and Guo, Cong},
  journal={Cognitive Computation},
  volume={17},
  number={2},
  pages={71},
url={https://doi.org/10.1007/s12559-025-10429-x},
  year={2025},
  publisher={Springer}
}

@inproceedings{bhattacharyya2025do,
title={{Do Machines Think Emotionally? A Cognitive Appraisal Analysis of Large Language Models}},
author={Sree Bhattacharyya and Lucas Craig and Tharun Dilliraj and Jia Li and James Z. Wang},
booktitle={Women in Machine Learning Workshop @ NeurIPS 2025},
year={2025},
url={https://openreview.net/forum?id=yaSopNnAby}
}

@inproceedings{bi-etal-2025-magi,
    title = {{{MAGI}: Multi-Agent Guided Interview for Psychiatric Assessment}},
    author = "Bi, Guanqun  and
      Chen, Zhuang  and
      Liu, Zhoufu  and
      Wang, Hongkai  and
      Xiao, Xiyao  and
      Xie, Yuqiang  and
      Zhang, Wen  and
      Huang, Yongkang  and
      Chen, Yuxuan  and
      Peng, Libiao  and
      Huang, Minlie",
    booktitle = "Findings of the Association for Computational Linguistics: ACL 2025",
    month = jul,
    year = "2025",
    address = "Vienna, Austria",
    publisher = "Association for Computational Linguistics",
    url = "https://aclanthology.org/2025.findings-acl.1278/",
    pages = "24898--24921",
    ISBN = "979-8-89176-256-5",
}

@article{croissant2024appraisal,
  title={{An appraisal-based chain-of-emotion architecture for affective language model game agents}},
  author={Croissant, Maximilian and Frister, Madeleine and Schofield, Guy and McCall, Cade},
  journal={Plos one},
  volume={19},
  number={5},
url={https://doi.org/10.1371/journal.pone.0301033},
  pages={e0301033},
  year={2024},
  publisher={Public Library of Science San Francisco, CA USA}
}

@misc{li2024enhancing,
  title={{Enhancing emotional generation capability of large language models via emotional chain-of-thought}},
  author={Li, Zaijing and Chen, Gongwei and Shao, Rui and Xie, Yuquan and Jiang, Dongmei and Nie, Liqiang},
url={https://arxiv.org/abs/2401.06836},
  year={2024}
}

@misc{openai2025gptoss120bgptoss20bmodel,
      title={gpt-oss-120b \& gpt-oss-20b Model Card}, 
      author={Sandhini Agarwal and Lama Ahmad and Jason Ai and Sam Altman and Andy Applebaum and Edwin Arbus and Rahul K. Arora and Yu Bai and Bowen Baker and Haiming Bao and Boaz Barak and Ally Bennett and Tyler Bertao and Nivedita Brett and Eugene Brevdo and Greg Brockman and Sebastien Bubeck and Che Chang and Kai Chen and Mark Chen and Enoch Cheung and Aidan Clark and Dan Cook and Marat Dukhan and Casey Dvorak and Kevin Fives and Vlad Fomenko and Timur Garipov and Kristian Georgiev and Mia Glaese and Tarun Gogineni and Adam Goucher and Lukas Gross and Katia Gil Guzman and John Hallman and Jackie Hehir and Johannes Heidecke and Alec Helyar and Haitang Hu and Romain Huet and Jacob Huh and Saachi Jain and Zach Johnson and Chris Koch and Irina Kofman and Dominik Kundel and Jason Kwon and Volodymyr Kyrylov and Elaine Ya Le and Guillaume Leclerc and James Park Lennon and Scott Lessans and Mario Lezcano-Casado and Yuanzhi Li and Zhuohan Li and Ji Lin and Jordan Liss and Lily and Liu and Jiancheng Liu and Kevin Lu and Chris Lu and Zoran Martinovic and Lindsay McCallum and Josh McGrath and Scott McKinney and Aidan McLaughlin and Song Mei and Steve Mostovoy and Tong Mu and Gideon Myles and Alexander Neitz and Alex Nichol and Jakub Pachocki and Alex Paino and Dana Palmie and Ashley Pantuliano and Giambattista Parascandolo and Jongsoo Park and Leher Pathak and Carolina Paz and Ludovic Peran and Dmitry Pimenov and Michelle Pokrass and Elizabeth Proehl and Huida Qiu and Gaby Raila and Filippo Raso and Hongyu Ren and Kimmy Richardson and David Robinson and Bob Rotsted and Hadi Salman and Suvansh Sanjeev and Max Schwarzer and D. Sculley and Harshit Sikchi and Kendal Simon and Karan Singhal and Yang Song and Dane Stuckey and Zhiqing Sun and Philippe Tillet and Sam Toizer and Foivos Tsimpourlas and Nikhil Vyas and Eric Wallace and Xin Wang and Miles Wang and Olivia Watkins and Kevin Weil and Amy Wendling and Kevin Whinnery and Cedric Whitney and Hannah Wong and Lin Yang and Yu Yang and Michihiro Yasunaga and Kristen Ying and Wojciech Zaremba and Wenting Zhan and Cyril Zhang and Brian Zhang and Eddie Zhang and Shengjia Zhao},
      year={2025},
      url={https://arxiv.org/abs/2508.10925}, 
}

@article{movassagh2023artificial,
  title={Artificial neural networks training algorithm integrating invasive weed optimization with differential evolutionary model},
  author={Movassagh, Ali Akbar and Alzubi, Jafar A and Gheisari, Mehdi and Rahimi, Mohamadtaghi and Mohan, Senthilkumar and Abbasi, Aaqif Afzaal and Nabipour, Narjes},
  journal={Journal of Ambient Intelligence and Humanized Computing},
  volume={14},
  number={5},
  pages={6017--6025},
  year={2023},
  publisher={Springer},
  url={https://link.springer.com/article/10.1007/s12652-020-02623-6}
}

@misc{singh2025openaigpt5card,
      title={OpenAI GPT-5 System Card}, 
      author={Aaditya Singh and Adam Fry and Adam Perelman and Adam Tart and Adi Ganesh and Ahmed El-Kishky and Aidan McLaughlin and Aiden Low and AJ Ostrow and Akhila Ananthram and Akshay Nathan and Alan Luo and Alec Helyar and Aleksander Madry and Aleksandr Efremov and Aleksandra Spyra and Alex Baker-Whitcomb and Alex Beutel and Alex Karpenko and Alex Makelov and Alex Neitz and Alex Wei and Alexandra Barr and Alexandre Kirchmeyer and Alexey Ivanov and Alexi Christakis and Alistair Gillespie and Allison Tam and Ally Bennett and Alvin Wan and Alyssa Huang and Amy McDonald Sandjideh and Amy Yang and Ananya Kumar and Andre Saraiva and Andrea Vallone and Andrei Gheorghe and Andres Garcia Garcia and Andrew Braunstein and Andrew Liu and Andrew Schmidt and Andrey Mereskin and Andrey Mishchenko and Andy Applebaum and Andy Rogerson and Ann Rajan and Annie Wei and Anoop Kotha and Anubha Srivastava and Anushree Agrawal and Arun Vijayvergiya and Ashley Tyra and Ashvin Nair and Avi Nayak and Ben Eggers and Bessie Ji and Beth Hoover and Bill Chen and Blair Chen and Boaz Barak and Borys Minaiev and Botao Hao and Bowen Baker and Brad Lightcap and Brandon McKinzie and Brandon Wang and Brendan Quinn and Brian Fioca and Brian Hsu and Brian Yang and Brian Yu and Brian Zhang and Brittany Brenner and Callie Riggins Zetino and Cameron Raymond and Camillo Lugaresi and Carolina Paz and Cary Hudson and Cedric Whitney and Chak Li and Charles Chen and Charlotte Cole and Chelsea Voss and Chen Ding and Chen Shen and Chengdu Huang and Chris Colby and Chris Hallacy and Chris Koch and Chris Lu and Christina Kaplan and Christina Kim and CJ Minott-Henriques and Cliff Frey and Cody Yu and Coley Czarnecki and Colin Reid and Colin Wei and Cory Decareaux and Cristina Scheau and Cyril Zhang and Cyrus Forbes and Da Tang and Dakota Goldberg and Dan Roberts and Dana Palmie and Daniel Kappler and Daniel Levine and Daniel Wright and Dave Leo and David Lin and David Robinson and Declan Grabb and Derek Chen and Derek Lim and Derek Salama and Dibya Bhattacharjee and Dimitris Tsipras and Dinghua Li and Dingli Yu and DJ Strouse and Drew Williams and Dylan Hunn and Ed Bayes and Edwin Arbus and Ekin Akyurek and Elaine Ya Le and Elana Widmann and Eli Yani and Elizabeth Proehl and Enis Sert and Enoch Cheung and Eri Schwartz and Eric Han and Eric Jiang and Eric Mitchell and Eric Sigler and Eric Wallace and Erik Ritter and Erin Kavanaugh and Evan Mays and Evgenii Nikishin and Fangyuan Li and Felipe Petroski Such and Filipe de Avila Belbute Peres and Filippo Raso and Florent Bekerman and Foivos Tsimpourlas and Fotis Chantzis and Francis Song and Francis Zhang and Gaby Raila and Garrett McGrath and Gary Briggs and Gary Yang and Giambattista Parascandolo and Gildas Chabot and Grace Kim and Grace Zhao and Gregory Valiant and Guillaume Leclerc and Hadi Salman and Hanson Wang and Hao Sheng and Haoming Jiang and Haoyu Wang and Haozhun Jin and Harshit Sikchi and Heather Schmidt and Henry Aspegren and Honglin Chen and Huida Qiu and Hunter Lightman and Ian Covert and Ian Kivlichan and Ian Silber and Ian Sohl and Ibrahim Hammoud and Ignasi Clavera and Ikai Lan and Ilge Akkaya and Ilya Kostrikov and Irina Kofman and Isak Etinger and Ishaan Singal and Jackie Hehir and Jacob Huh and Jacqueline Pan and Jake Wilczynski and Jakub Pachocki and James Lee and James Quinn and Jamie Kiros and Janvi Kalra and Jasmyn Samaroo and Jason Wang and Jason Wolfe and Jay Chen and Jay Wang and Jean Harb and Jeffrey Han and Jeffrey Wang and Jennifer Zhao and Jeremy Chen and Jerene Yang and Jerry Tworek and Jesse Chand and Jessica Landon and Jessica Liang and Ji Lin and Jiancheng Liu and Jianfeng Wang and Jie Tang and Jihan Yin and Joanne Jang and Joel Morris and Joey Flynn and Johannes Ferstad and Johannes Heidecke and John Fishbein and John Hallman and Jonah Grant and Jonathan Chien and Jonathan Gordon and Jongsoo Park and Jordan Liss and Jos Kraaijeveld and Joseph Guay and Joseph Mo and Josh Lawson and Josh McGrath and Joshua Vendrow and Joy Jiao and Julian Lee and Julie Steele and Julie Wang and Junhua Mao and Kai Chen and Kai Hayashi and Kai Xiao and Kamyar Salahi and Kan Wu and Karan Sekhri and Karan Sharma and Karan Singhal and Karen Li and Kenny Nguyen and Keren Gu-Lemberg and Kevin King and Kevin Liu and Kevin Stone and Kevin Yu and Kristen Ying and Kristian Georgiev and Kristie Lim and Kushal Tirumala and Kyle Miller and Lama Ahmad and Larry Lv and Laura Clare and Laurance Fauconnet and Lauren Itow and Lauren Yang and Laurentia Romaniuk and Leah Anise and Lee Byron and Leher Pathak and Leon Maksin and Leyan Lo and Leyton Ho and Li Jing and Liang Wu and Liang Xiong and Lien Mamitsuka and Lin Yang and Lindsay McCallum and Lindsey Held and Liz Bourgeois and Logan Engstrom and Lorenz Kuhn and Louis Feuvrier and Lu Zhang and Lucas Switzer and Lukas Kondraciuk and Lukasz Kaiser and Manas Joglekar and Mandeep Singh and Mandip Shah and Manuka Stratta and Marcus Williams and Mark Chen and Mark Sun and Marselus Cayton and Martin Li and Marvin Zhang and Marwan Aljubeh and Matt Nichols and Matthew Haines and Max Schwarzer and Mayank Gupta and Meghan Shah and Melody Huang and Meng Dong and Mengqing Wang and Mia Glaese and Micah Carroll and Michael Lampe and Michael Malek and Michael Sharman and Michael Zhang and Michele Wang and Michelle Pokrass and Mihai Florian and Mikhail Pavlov and Miles Wang and Ming Chen and Mingxuan Wang and Minnia Feng and Mo Bavarian and Molly Lin and Moose Abdool and Mostafa Rohaninejad and Nacho Soto and Natalie Staudacher and Natan LaFontaine and Nathan Marwell and Nelson Liu and Nick Preston and Nick Turley and Nicklas Ansman and Nicole Blades and Nikil Pancha and Nikita Mikhaylin and Niko Felix and Nikunj Handa and Nishant Rai and Nitish Keskar and Noam Brown and Ofir Nachum and Oleg Boiko and Oleg Murk and Olivia Watkins and Oona Gleeson and Pamela Mishkin and Patryk Lesiewicz and Paul Baltescu and Pavel Belov and Peter Zhokhov and Philip Pronin and Phillip Guo and Phoebe Thacker and Qi Liu and Qiming Yuan and Qinghua Liu and Rachel Dias and Rachel Puckett and Rahul Arora and Ravi Teja Mullapudi and Raz Gaon and Reah Miyara and Rennie Song and Rishabh Aggarwal and RJ Marsan and Robel Yemiru and Robert Xiong and Rohan Kshirsagar and Rohan Nuttall and Roman Tsiupa and Ronen Eldan and Rose Wang and Roshan James and Roy Ziv and Rui Shu and Ruslan Nigmatullin and Saachi Jain and Saam Talaie and Sam Altman and Sam Arnesen and Sam Toizer and Sam Toyer and Samuel Miserendino and Sandhini Agarwal and Sarah Yoo and Savannah Heon and Scott Ethersmith and Sean Grove and Sean Taylor and Sebastien Bubeck and Sever Banesiu and Shaokyi Amdo and Shengjia Zhao and Sherwin Wu and Shibani Santurkar and Shiyu Zhao and Shraman Ray Chaudhuri and Shreyas Krishnaswamy and Shuaiqi and Xia and Shuyang Cheng and Shyamal Anadkat and Simón Posada Fishman and Simon Tobin and Siyuan Fu and Somay Jain and Song Mei and Sonya Egoian and Spencer Kim and Spug Golden and SQ Mah and Steph Lin and Stephen Imm and Steve Sharpe and Steve Yadlowsky and Sulman Choudhry and Sungwon Eum and Suvansh Sanjeev and Tabarak Khan and Tal Stramer and Tao Wang and Tao Xin and Tarun Gogineni and Taya Christianson and Ted Sanders and Tejal Patwardhan and Thomas Degry and Thomas Shadwell and Tianfu Fu and Tianshi Gao and Timur Garipov and Tina Sriskandarajah and Toki Sherbakov and Tomer Kaftan and Tomo Hiratsuka and Tongzhou Wang and Tony Song and Tony Zhao and Troy Peterson and Val Kharitonov and Victoria Chernova and Vineet Kosaraju and Vishal Kuo and Vitchyr Pong and Vivek Verma and Vlad Petrov and Wanning Jiang and Weixing Zhang and Wenda Zhou and Wenlei Xie and Wenting Zhan and Wes McCabe and Will DePue and Will Ellsworth and Wulfie Bain and Wyatt Thompson and Xiangning Chen and Xiangyu Qi and Xin Xiang and Xinwei Shi and Yann Dubois and Yaodong Yu and Yara Khakbaz and Yifan Wu and Yilei Qian and Yin Tat Lee and Yinbo Chen and Yizhen Zhang and Yizhong Xiong and Yonglong Tian and Young Cha and Yu Bai and Yu Yang and Yuan Yuan and Yuanzhi Li and Yufeng Zhang and Yuguang Yang and Yujia Jin and Yun Jiang and Yunyun Wang and Yushi Wang and Yutian Liu and Zach Stubenvoll and Zehao Dou and Zheng Wu and Zhigang Wang},
      year={2025},
      url={https://arxiv.org/abs/2601.03267}, 
}

@inproceedings{10.1145/3544548.3580688,
author = {H\"{a}m\"{a}l\"{a}inen, Perttu and Tavast, Mikke and Kunnari, Anton},
title = {{Evaluating Large Language Models in Generating Synthetic HCI Research Data: a Case Study}},
year = {2023},
isbn = {9781450394215},
publisher = {Association for Computing Machinery},
address = {New York, NY, USA},
url = {https://doi.org/10.1145/3544548.3580688},
booktitle = {Proceedings of the 2023 CHI Conference on Human Factors in Computing Systems},
articleno = {433},
numpages = {19},
location = {Hamburg, Germany},
series = {CHI '23}
}

@article{10.1145/3774896,
author = {Plaat, Aske and Wong, Annie and Verberne, Suzan and Broekens, Joost and Van Stein, Niki and B\"{a}ck, Thomas},
title = {{Multi-Step Reasoning with Large Language Models, a Survey}},
year = {2025},
issue_date = {April 2026},
publisher = {Association for Computing Machinery},
address = {New York, NY, USA},
volume = {58},
number = {6},
issn = {0360-0300},
url = {https://doi.org/10.1145/3774896},
journal = {ACM Comput. Surv.},
month = dec,
articleno = {160},
numpages = {35}
}

@article{JIM2024100059,
title = {{Recent advancements and challenges of NLP-based sentiment analysis: A state-of-the-art review}},
journal = {Natural Language Processing Journal},
volume = {6},
pages = {100059},
year = {2024},
issn = {2949-7191},
doi = {https://doi.org/10.1016/j.nlp.2024.100059},
author = {Jamin Rahman Jim and Md Apon Riaz Talukder and Partha Malakar and Md Mohsin Kabir and Kamruddin Nur and M.F. Mridha},
}

@misc{grattafiori2024llama3herdmodels,
      title={{The Llama 3 Herd of Models}}, 
      author={Aaron Grattafiori and Abhimanyu Dubey and Abhinav Jauhri and Abhinav Pandey and Abhishek Kadian and Ahmad Al-Dahle and Aiesha Letman and Akhil Mathur and Alan Schelten and Alex Vaughan and Amy Yang and Angela Fan and Anirudh Goyal and Anthony Hartshorn and Aobo Yang and othere},
      year={2024},
      url={https://arxiv.org/abs/2407.21783}, 
}

@misc{mistral2024large,
  title={{Mistral Large 2}},
  author={MistralAI},
  year={2024},
  url={https://mistral.ai/news/mistral-large-2407}
}

@misc{gemmateam2025gemma3technicalreport,
      title={{Gemma 3 Technical Report}}, 
      author={Gemma Team: Aishwarya Kamath and Johan Ferret and Shreya Pathak and Nino Vieillard and Ramona Merhej and Sarah Perrin and Tatiana Matejovicova and Alexandre Ramé and Morgane Rivière and Louis Rouillard and Thomas Mesnard and Geoffrey Cideron and Jean-bastien Grill and Sabela Ramos and Edouard Yvinec and Michelle Casbon and Etienne Pot and Ivo Penchev and Gaël Liu and Francesco Visin and Kathleen Kenealy and Lucas Beyer and Xiaohai Zhai and Anton Tsitsulin and Robert Busa-Fekete and Alex Feng and Noveen Sachdeva and Benjamin Coleman and Yi Gao and Basil Mustafa and Iain Barr and Emilio Parisotto and David Tian and Matan Eyal and Colin Cherry and Jan-Thorsten Peter and Danila Sinopalnikov and Surya Bhupatiraju and Rishabh Agarwal and Mehran Kazemi and Dan Malkin and Ravin Kumar and David Vilar and Idan Brusilovsky and Jiaming Luo and Andreas Steiner and Abe Friesen and Abhanshu Sharma and Abheesht Sharma and Adi Mayrav Gilady and Adrian Goedeckemeyer and Alaa Saade and Alex Feng and Alexander Kolesnikov and Alexei Bendebury and Alvin Abdagic and Amit Vadi and András György and André Susano Pinto and Anil Das and Ankur Bapna and Antoine Miech and Antoine Yang and Antonia Paterson and Ashish Shenoy and Ayan Chakrabarti and Bilal Piot and Bo Wu and Bobak Shahriari and Bryce Petrini and Charlie Chen and Charline Le Lan and Christopher A. Choquette-Choo and CJ Carey and Cormac Brick and Daniel Deutsch and Danielle Eisenbud and Dee Cattle and Derek Cheng and Dimitris Paparas and Divyashree Shivakumar Sreepathihalli and Doug Reid and Dustin Tran and Dustin Zelle and Eric Noland and Erwin Huizenga and Eugene Kharitonov and Frederick Liu and Gagik Amirkhanyan and Glenn Cameron and Hadi Hashemi and Hanna Klimczak-Plucińska and Harman Singh and Harsh Mehta and Harshal Tushar Lehri and Hussein Hazimeh and Ian Ballantyne and Idan Szpektor and Ivan Nardini and Jean Pouget-Abadie and Jetha Chan and Joe Stanton and John Wieting and Jonathan Lai and Jordi Orbay and Joseph Fernandez and Josh Newlan and Ju-yeong Ji and Jyotinder Singh and Kat Black and Kathy Yu and Kevin Hui and Kiran Vodrahalli and Klaus Greff and Linhai Qiu and Marcella Valentine and Marina Coelho and Marvin Ritter and Matt Hoffman and Matthew Watson and Mayank Chaturvedi and Michael Moynihan and Min Ma and Nabila Babar and Natasha Noy and Nathan Byrd and Nick Roy and Nikola Momchev and Nilay Chauhan and Noveen Sachdeva and Oskar Bunyan and Pankil Botarda and Paul Caron and Paul Kishan Rubenstein and Phil Culliton and Philipp Schmid and Pier Giuseppe Sessa and Pingmei Xu and Piotr Stanczyk and Pouya Tafti and Rakesh Shivanna and Renjie Wu and Renke Pan and Reza Rokni and Rob Willoughby and Rohith Vallu and Ryan Mullins and Sammy Jerome and Sara Smoot and Sertan Girgin and Shariq Iqbal and Shashir Reddy and Shruti Sheth and Siim Põder and Sijal Bhatnagar and Sindhu Raghuram Panyam and Sivan Eiger and Susan Zhang and Tianqi Liu and Trevor Yacovone and Tyler Liechty and Uday Kalra and Utku Evci and Vedant Misra and Vincent Roseberry and Vlad Feinberg and Vlad Kolesnikov and Woohyun Han and Woosuk Kwon and Xi Chen and Yinlam Chow and Yuvein Zhu and Zichuan Wei and Zoltan Egyed and Victor Cotruta and Minh Giang and Phoebe Kirk and Anand Rao and Kat Black and Nabila Babar and Jessica Lo and Erica Moreira and Luiz Gustavo Martins and Omar Sanseviero and Lucas Gonzalez and Zach Gleicher and Tris Warkentin and Vahab Mirrokni and Evan Senter and Eli Collins and Joelle Barral and Zoubin Ghahramani and Raia Hadsell and Yossi Matias and D. Sculley and Slav Petrov and Noah Fiedel and Noam Shazeer and Oriol Vinyals and Jeff Dean and Demis Hassabis and Koray Kavukcuoglu and Clement Farabet and Elena Buchatskaya and Jean-Baptiste Alayrac and Rohan Anil and Dmitry and Lepikhin and Sebastian Borgeaud and Olivier Bachem and Armand Joulin and Alek Andreev and Cassidy Hardin and Robert Dadashi and Léonard Hussenot},
      year={2025},
      url={https://arxiv.org/abs/2503.19786}, 
}

@article{SUPRIYONO2024100173,
title = {{Advancements in natural language processing: Implications, challenges, and future directions}},
journal = {Telematics and Informatics Reports},
volume = {16},
pages = {100173},
year = {2024},
issn = {2772-5030},
doi = {https://doi.org/10.1016/j.teler.2024.100173},
author = { Supriyono and Aji Prasetya Wibawa and  Suyono and Fachrul Kurniawan},
}

@article{ekman1992argument,
  title={{An argument for basic emotions}},
  author={Ekman, Paul},
  journal={Cognition \& emotion},
  volume={6},
  number={3-4},
  pages={169--200},
url={https://doi.org/10.1080/02699939208411068},
  year={1992},
  publisher={Taylor \& Francis}
}

@inproceedings{Barei-2024, 
series={WWW ’24},
   title={{English Prompts are Better for NLI-based Zero-Shot Emotion Classification than Target-Language Prompts}},
   url={http://dx.doi.org/10.1145/3589335.3651902},
   booktitle={Companion Proceedings of the ACM Web Conference 2024},
   publisher={ACM},
   author={Bareiß, Patrick and Klinger, Roman and Barnes, Jeremy},
   year={2024},
   month=may, pages={1318–1326},
   collection={WWW ’24} }

@inproceedings{muhammad-etal-2025-semeval,
    title = {{{S}em{E}val-2025 Task 11: Bridging the Gap in Text-Based Emotion Detection}},
    author = "Muhammad, Shamsuddeen Hassan  and
      Ousidhoum, Nedjma  and
      Abdulmumin, Idris  and
      Yimam, Seid Muhie  and
      Wahle, Jan Philip  and
      Lima Ruas, Terry  and
      Beloucif, Meriem  and
      De Kock, Christine  and
      Belay, Tadesse Destaw  and
      Ahmad, Ibrahim Said  and
      Surange, Nirmal  and
      Teodorescu, Daniela  and
      Adelani, David Ifeoluwa  and
      Aji, Alham Fikri  and
      Ali, Felermino Dario Mario  and
      Araujo, Vladimir  and
      Ayele, Abinew Ali  and
      Ignat, Oana  and
      Panchenko, Alexander  and
      Zhou, Yi  and
      Mohammad, Saif",
    booktitle = "Proceedings of the 19th International Workshop on Semantic Evaluation (SemEval-2025)",
    month = jul,
    year = "2025",
    address = "Vienna, Austria",
    publisher = "Association for Computational Linguistics",
    url = "https://aclanthology.org/2025.semeval-1.327/",
    pages = "2558--2569",
    doi="",
    ISBN = "979-8-89176-273-2"
}

@article{wei2022chain,
  title={{Chain-of-thought prompting elicits reasoning in large language models}},
  author={Wei, Jason and Wang, Xuezhi and Schuurmans, Dale and Bosma, Maarten and Xia, Fei and Chi, Ed and Le, Quoc V and Zhou, Denny and others},
  journal={Advances in neural information processing systems},
  volume={35},
url={https://dl.acm.org/doi/10.5555/3600270.3602070},
  pages={24824--24837},
  year={2022}
}

@misc{li2024ECOT,
      title={{Enhancing Emotional Generation Capability of Large Language Models via Emotional Chain-of-Thought}}, 
      author={Zaijing Li and Gongwei Chen and Rui Shao and Yuquan Xie and Dongmei Jiang and Liqiang Nie},
      year={2024},
      eprint={2401.06836},
      archivePrefix={arXiv},
      primaryClass={cs.CL},
      url={https://arxiv.org/abs/2401.06836}, 
}

@misc{li2023large,
      title={{Large Language Models Understand and Can be Enhanced by Emotional Stimuli}}, 
      author={Cheng Li and Jindong Wang and Yixuan Zhang and Kaijie Zhu and Wenxin Hou and Jianxun Lian and Fang Luo and Qiang Yang and Xing Xie},
      year={2023},
      primaryClass={cs.CL},
      url={https://arxiv.org/abs/2307.11760}, 
}

@article{DeepSeek-r1,
   title={DeepSeek-R1 incentivizes reasoning in LLMs through reinforcement learning},
   volume={645},
   ISSN={1476-4687},
   url={http://dx.doi.org/10.1038/s41586-025-09422-z},
   number={8081},
   journal={Nature},
   publisher={Springer Science and Business Media LLC},
   author={Guo, Daya and Yang, Dejian and Zhang, Haowei and Song, Junxiao and Wang, Peiyi and Zhu, Qihao and Xu, Runxin and Zhang, Ruoyu and Ma, Shirong and Bi, Xiao and Zhang, Xiaokang and Yu, Xingkai and Wu, Yu and Wu, Z. F. and Gou, Zhibin and Shao, Zhihong and Li, Zhuoshu and Gao, Ziyi and Liu, Aixin and Xue, Bing and Wang, Bingxuan and Wu, Bochao and Feng, Bei and Lu, Chengda and Zhao, Chenggang and Deng, Chengqi and Ruan, Chong and Dai, Damai and Chen, Deli and Ji, Dongjie and Li, Erhang and Lin, Fangyun and Dai, Fucong and Luo, Fuli and Hao, Guangbo and Chen, Guanting and Li, Guowei and Zhang, H. and Xu, Hanwei and Ding, Honghui and Gao, Huazuo and Qu, Hui and Li, Hui and Guo, Jianzhong and Li, Jiashi and Chen, Jingchang and Yuan, Jingyang and Tu, Jinhao and Qiu, Junjie and Li, Junlong and Cai, J. L. and Ni, Jiaqi and Liang, Jian and Chen, Jin and Dong, Kai and Hu, Kai and You, Kaichao and Gao, Kaige and Guan, Kang and Huang, Kexin and Yu, Kuai and Wang, Lean and Zhang, Lecong and Zhao, Liang and Wang, Litong and Zhang, Liyue and Xu, Lei and Xia, Leyi and Zhang, Mingchuan and Zhang, Minghua and Tang, Minghui and Zhou, Mingxu and Li, Meng and Wang, Miaojun and Li, Mingming and Tian, Ning and Huang, Panpan and Zhang, Peng and Wang, Qiancheng and Chen, Qinyu and Du, Qiushi and Ge, Ruiqi and Zhang, Ruisong and Pan, Ruizhe and Wang, Runji and Chen, R. J. and Jin, R. L. and Chen, Ruyi and Lu, Shanghao and Zhou, Shangyan and Chen, Shanhuang and Ye, Shengfeng and Wang, Shiyu and Yu, Shuiping and Zhou, Shunfeng and Pan, Shuting and Li, S. S. and Zhou, Shuang and Wu, Shaoqing and Yun, Tao and Pei, Tian and Sun, Tianyu and Wang, T. and Zeng, Wangding and Liu, Wen and Liang, Wenfeng and Gao, Wenjun and Yu, Wenqin and Zhang, Wentao and Xiao, W. L. and An, Wei and Liu, Xiaodong and Wang, Xiaohan and Chen, Xiaokang and Nie, Xiaotao and Cheng, Xin and Liu, Xin and Xie, Xin and Liu, Xingchao and Yang, Xinyu and Li, Xinyuan and Su, Xuecheng and Lin, Xuheng and Li, X. Q. and Jin, Xiangyue and Shen, Xiaojin and Chen, Xiaosha and Sun, Xiaowen and Wang, Xiaoxiang and Song, Xinnan and Zhou, Xinyi and Wang, Xianzu and Shan, Xinxia and Li, Y. K. and Wang, Y. Q. and Wei, Y. X. and Zhang, Yang and Xu, Yanhong and Li, Yao and Zhao, Yao and Sun, Yaofeng and Wang, Yaohui and Yu, Yi and Zhang, Yichao and Shi, Yifan and Xiong, Yiliang and He, Ying and Piao, Yishi and Wang, Yisong and Tan, Yixuan and Ma, Yiyang and Liu, Yiyuan and Guo, Yongqiang and Ou, Yuan and Wang, Yuduan and Gong, Yue and Zou, Yuheng and He, Yujia and Xiong, Yunfan and Luo, Yuxiang and You, Yuxiang and Liu, Yuxuan and Zhou, Yuyang and Zhu, Y. X. and Huang, Yanping and Li, Yaohui and Zheng, Yi and Zhu, Yuchen and Ma, Yunxian and Tang, Ying and Zha, Yukun and Yan, Yuting and Ren, Z. Z. and Ren, Zehui and Sha, Zhangli and Fu, Zhe and Xu, Zhean and Xie, Zhenda and Zhang, Zhengyan and Hao, Zhewen and Ma, Zhicheng and Yan, Zhigang and Wu, Zhiyu and Gu, Zihui and Zhu, Zijia and Liu, Zijun and Li, Zilin and Xie, Ziwei and Song, Ziyang and Pan, Zizheng and Huang, Zhen and Xu, Zhipeng and Zhang, Zhongyu and Zhang, Zhen},
   year={2025},
   month=sep, pages={633–638} 
}

@incollection{appraisal-Moors2017,
  title = {{Appraisal Theory of Emotion}},
  booktitle = {Encyclopedia of Personality and Individual Differences},
  author = {Moors, Agnes},
  year = {2017},
  pages = {1--9},
  publisher = {Springer International Publishing},
  location = {Cham},
  url = {https://doi.org/10.1007/978-3-319-28099-8_493-1},
  isbn = {978-3-319-28099-8}
}

\end{document}